\documentclass[acmsmall,screen]{acmart}
\usepackage{CJKutf8}
\usepackage[utf8]{inputenc}
\usepackage[T1]{fontenc}
\usepackage{textcomp}
\usepackage{xcolor,colortbl}
\usepackage[most]{tcolorbox}
\usepackage{multirow}
\usepackage{multicol}
\usepackage{threeparttable}
\usepackage{soul}
\usepackage{hyperref} 
\usepackage{longtable}
\usepackage{subfig}
\usepackage{overpic}
\usepackage{array}
\usepackage{float}
\usepackage{tabularx}
\usepackage[english]{babel}
\usepackage{booktabs}
\usepackage{hyphenat}

\AtBeginDocument{
  \providecommand\BibTeX{{
    \normalfont B\kern-0.5em{\scshape i\kern-0.25em b}\kern-0.8em\TeX}}}

\newif\ifdraft
\drafttrue

\renewcommand\footnotetextcopyrightpermission[1]{} 
\begin{document}
\setcopyright{acmcopyright}
\copyrightyear{2026}
\acmYear{2026}
\acmDOI{XX.XXXX/XXXXXXX.XXXXXXX}

\acmJournal{JACM}
\acmVolume{1}
\acmNumber{1}
\acmArticle{1}
\acmMonth{1}

\title[A User-driven Design Framework for Robotaxis]{A User-driven Design Framework for Robotaxis}

\author{Yue Deng}
\affiliation{%
  \institution{Hong Kong University of Science and Technology}
  \city{Hong Kong}
  \country{China}
}
\email{ydengbi@connect.ust.hk}

\author{Changyang He}
\affiliation{%
  \institution{Max Planck Institute for Security and Privacy}
  \city{Bochum}
  \country{Germany}
}
\email{changyang.he.young@gmail.com}

\renewcommand{\shortauthors}{Deng et al.}

\newcommand{\gr}{\leavevmode\color{lightgray}}
\newcommand{\cgr}{\cellcolor{green!25}}
\newcommand{\crd}{\cellcolor{pink!25}}

\begin{abstract}

Robotaxis are emerging as a promising form of urban mobility, but removing human drivers fundamentally reshapes passenger-vehicle interaction and raises new design challenges. To inform robotaxi design based on real-world experience, we conducted 18 semi-structured interviews and autoethnographic ride experiences to examine users' perceptions, experiences, and expectations for robotaxi design. We found that users valued benefits such as increased agency and consistent driving. However, they also encountered challenges such as limited flexibility, insufficient transparency, and emergency handling concerns. Notably, users perceived robotaxis not merely as a mode of transportation, but as autonomous, semi-private transitional spaces, which made users feel less socially intrusive to engage in personal activities. Safety perceptions were polarized: some felt anxiety about reduced control, while others viewed robotaxis as safer than humans due to their cautious, law-abiding nature. Based on the findings, we propose a user-driven design framework spanning hailing, pick-up, traveling, and drop-off phases to support trustworthy, transparent, and accountable robotaxi design.

\end{abstract}

\ccsdesc[500]{Human-centered computing~Empirical studies in interaction design}
\ccsdesc[500]{Human-centered computing~Empirical studies in collaborative and social computing}
\ccsdesc[500]{Human-centered computing~Interaction paradigms}
\ccsdesc[500]{Human-centered computing~Empirical studies in HCI}

\keywords{autonomous driving, robotaxi, design framework}

\maketitle

\section{INTRODUCTION}

With the rapid advancement of autonomous driving technologies, \textit{robotaxis}, autonomous vehicles operated for ride-hailing services, have demonstrated increasing potential for future urban mobility~\cite{hancock2019future,pavone2016autonomous}. By eliminating the need for human drivers, robotaxis can significantly reduce labor costs while also offering potential benefits such as reduced traffic congestion and lower carbon emissions~\cite{zhang2024evolution}. As of January 2026, robotaxi services have been deployed in pilot programs across several countries such as the United States~\cite{waymo}, China~\cite{ponyai,ApolloGo,WeRide}, and the United Arab Emirates~\cite{UberUAE}. For example, Apollo Go launched autonomous mobility services in several cities in China in 2022, and has accumulated 17 million rides by November 2025~\cite{apollogoReport}. As widespread deployment of robotaxis becomes more plausible, governments around the world have also started developing regulatory frameworks to govern their operation~\cite{CalAV,GermanyAV,ukAVAct}.

Although autonomous driving technologies are continuously advancing toward (and sometimes beyond) human-level driving performance, existing research has focused predominantly on improving driving capabilities~\cite{yurtsever2020survey,zhao2025survey,rajaram2022development,wu2025enhancing,qi2024multi}, while largely overlooking the novel interaction paradigms introduced by the replacement of human drivers with machines. This shift not only alters passengers' riding experience due to changes in driving capability and style, but also fundamentally reshapes the interaction between passengers and vehicles~\cite{al2024navigating,brandebusemeyer2022travelers,kim2024timelytale}. In particular, the absence of a human driver redefines passengers' perceptions of privacy within the vehicle~\cite{al2024navigating}, profoundly influences their sense of safety~\cite{gu2024different} and trust~\cite{liu2025understanding} in the driving system, and even prompts a re-examination of driving-related ethical considerations~\cite{jia2024will}. As such, new design requirements of robotaxis emerge, such as explanations of vehicle decisions~\cite{hua2022establish}, enriched in-vehicle experiences~\cite{auhspace}, and accessible safety mechanisms that allow passengers to exit the vehicle in emergency situations~\cite{lim2021ux}. 


Since large-scale robotaxi deployment has yet to mature, existing HCI literature on robotaxi user experience mainly adopts simulation approaches such as virtual reality (VR)~\cite{brandt2026communicating} and Wizard of Oz (WOZ) studies~\cite{meurer2020wizard,lim2021ux,yoo2024anxiety,meister2025external,weschke2021asking}. Despite the value of these early-stage studies, research based on simulated or hypothetical perspectives has inherent limitations: for example, Wizard-of-Oz (WOZ) approaches cannot faithfully replicate real robotaxi driving patterns. To date, there has been little research investigating user experience in real-world autonomous driving contexts and leveraging such insights to inform the design of robotaxis. With the emergence of robotaxi pilot programs and the accumulation of real-world user experience data, this work presents the first effort to examine users' experiences, perceptions, and challenges when interacting with real-world robotaxi services, providing practice-driven insights for robotaxi system design.

Specifically, this paper seeks to address the following research questions:

\begin{itemize}
  
  \item \textbf{RQ1}: What factors motivate users to use robotaxis?
  \item \textbf{RQ2}: What advantages do users perceive in robotaxis, and what challenges or drawbacks do they identify?
  \item \textbf{RQ3}: How do users perceive privacy, safety, and ethical issues related to robotaxis, and how do these perceptions shape their trust?
  \item \textbf{RQ4}: What design features do users expect for future robotaxi systems?

\end{itemize}

To address the RQs outlined above, this work adopted a qualitative approach that integrates complementary perspectives from interviews and autoethnography. Drawing on 18 interviews with robotaxi users and 22 instances of first-person ride experiences conducted across three cities with robotaxi commercial programs in China, the study combined participants' reflective accounts with the researchers' situated, experiential observations, enabling a more comprehensive understanding of user experiences in real-world robotaxi contexts.

We found that initial robotaxi adoption was largely driven by low cost, recommendations from acquaintances, and curiosity toward the novelty of robotaxi (RQ1). After use, participants reported a set of distinctive advantages, such as a sense of agency both in perceived operational control and in psychological feelings of freedom and ownership, the perceived consistency of robotaxi behavior and standardized ride quality across trips, and the service's transparency around routing and pricing. At the same time, users encountered persistent challenges. One of the most salient frictions was limited flexibility across over-cautious and conservative driving and fixed pick-up and drop-off parking. Participants also described gaps in transparency (e.g., temporal, operational, and financial transparency), practical management issues (e.g., vehicle hygiene), concerns about system robustness (e.g., robustness in unusual or edge-case road conditions), and uncertainty about how to respond to emergencies (e.g., inadequate user education and support) as critical challenges of robotaxi adoption (RQ2).




Regarding privacy, our findings indicate that users perceived robotaxis as a semi-private space to engage in personal activities compared to driver-present vehicles, while remaining aware of ongoing in-vehicle monitoring and actively modulating their behaviors. Users tended to be indifferent rather than resistant to data collection in robotaxis and generally accepted it when it was framed as necessary or beneficial. Safety perceptions were polarized: some users emphasized anxiety stemming from factors such as reduced personal control and doubts about designers' real-world driving experience, whereas others viewed robotaxis as safer than human-driven vehicles due to factors such as law-abiding, cautious driving and the reduced risk of interpersonal conflict. Users held ethical concerns about robotaxis, such as accident accountability, responsibility in feedback attribution, discomfort with delegating life-and-death control, and the absence of human-like driving norms. Trust, finally, was shaped by personal, social, and technical factors, such as personal ride experience, regulation and institutional oversight, and vehicle safety-related design features (RQ3).

Finally, our findings captured user expectations for future robotaxi design, including control and trust expectations (e.g., explainable feedback and emergency support), experiential and affective expectations (e.g., in-ride entertainment and more seamless automation), and personalized and adaptable expectations (e.g., driving customization options and personalized companion services) (RQ4).

Our work makes the following contributions to user-centered robotaxi design:
\begin{itemize}
    \item We provide the first real-world account of robotaxi use that combines interview and autoethnographic perspectives, capturing users' motivations, practices, and perceptions in robotaxi systems.

    \item We reveal rich empirical insights into perceived privacy and agency in robotaxi use, emphasizing robotaxis not merely as transportation but as an autonomous, semi-private transitional space, which afford new design opportunities for shaping the meaning of ride experiences.

    \item We contribute a nuanced understanding of the interplay between perceived safety and user trust in robotaxis, highlighting how they are dynamically constructed through users' continuous sensemaking, explainability features, and shifting boundaries of human–machine autonomy and accountability.

    \item We derive a user-driven, end-to-end design framework for robotaxis that translates user-centered insights into actionable design implications, shedding light on trustworthy, transparent, and accountable robotaxi design.
\end{itemize}


\section{BACKGROUND AND RELATED WORK}

\subsection{The Development of Robotaxi: Technology, Regulations, and Deployment}

Driven by efforts from both academia and industry, autonomous driving has achieved significant performance improvements in perception, decision-making, and control in recent years~\cite{yurtsever2020survey,zhao2025survey}. Progress in autonomous driving has not only reduced the burden on private car drivers, but has also provided the technological foundation for robotaxis, turning autonomous mobility on demand (AMoD) from a sustainability-oriented vision into a practical mobility solution~\cite{pavone2016autonomous,hancock2019future}. Mainstream robotaxi companies (such as Waymo~\cite{waymo}, Pony.ai~\cite{ponyai}, and Apollo Go~\cite{ApolloGo}) rely on multi-sensor fusion, integrating LiDAR, cameras, and radar together with high-definition maps to improve perception robustness and safety through redundancy. Today, these leading robotaxi systems have largely reached SAE (Society of Automotive Engineers) Level 4 autonomous driving~\cite{on2021taxonomy}, i.e., the vehicle can operate fully autonomously under specific scenarios and conditions~\cite{liu2025understanding}. Nevertheless, robotaxis remain constrained in uncertain conditions such as complex traffic scenarios and adverse weather, and a substantial gap to SAE Level 5 autonomy (full autonomy in all environments) persists~\cite{fowler2024robotaxi}.


As the large-scale adoption of robotaxis becomes increasingly foreseeable, governments worldwide have begun to establish regulatory frameworks for robotaxi operations. In 2021, the Road Traffic Act Amendment in Germany allowed motor vehicles with autonomous driving capabilities (corresponding to SAE level 4 automation) in specified operating areas on public roads~\cite{GermanyAV}. The United Kingdom enables commercial robotaxi deployment through the Automated Vehicles Act 2024, establishing a national authorization and liability framework for self-driving passenger services~\cite{ukAVAct}. In the United States, the federal government oversees vehicle safety standards while individual states regulate testing, deployment, and commercial passenger services; notably, the California Public Utilities Commission approved the deployment permit for ``Driverless AV Passenger Service'' without a driver in the vehicle in 2022~\cite{CalAV}. Similarly, China authorizes robotaxis through city-level permits. For example, Shenzhen has granted Pony.ai the city's first citywide permit for fully driverless commercial robotaxi operations in 2025~\cite{ShenzhenAV}.


As of January 2026, major robotaxi companies worldwide have begun moving from pilot testing into early commercial operations. In the United States, Waymo operates fully driverless services in cities such as Phoenix, San Francisco, and Los Angeles~\cite{waymo}. In China, Baidu Apollo Go~\cite{ApolloGo}, Pony.ai~\cite{ponyai}, and WeRide~\cite{WeRide} run large-scale robotaxi programs across multiple major cities, including Wuhan, Chongqing, Shenzhen, and Guangzhou~\cite{ChinaRobotaxi}. In the Middle East, WeRide and Uber have operated fully driverless robotaxi services in Abu Dhabi, UAE, since 2024, expanding to over 100 robotaxis by the end of 2025~\cite{UberUAE}.

With the rapid development across technology, regulations, and deployment, robotaxi services have gradually scaled from limited-hour trials in small test zones to full-day, large-area operations providing paid autonomous rides to the public. However, user experience in real-world robotaxi services has not yet been systematically studied. This work seeks to address this gap, shedding light on user-friendly robotaxi design.

\subsection{Human Factors in Robotaxi Adoption and Use}

Given the rapid development of robotaxis, scholars have begun to explore human factors in robotaxi adoption and use. One line of research has focused on user acceptance models for robotaxis~\cite{liu2022factors,wei2024using,lee2022effect,yang2023does,yao2025transition,liu2020study,lin2025future,gu2024promote}. For example, Liu et al. adopted technology acceptance model (TAM) to understand user acceptance of robotaxi services, and found that both the typical TAM factors (perceived ease of use, perceived usefulness, and attitude) and external factors (trust, government support, social influence, and perceived enjoyment) significantly correlated with users' intention to use robotaxis~\cite{liu2022factors}. Wei et al. employed the Extended Unified Theory of Acceptance and Use of Technology (UTAUT2) to understand the factors that influence adoption and usage of robotaxis in China, and found that performance expectancy, hedonic motivation, and price value significantly influence users' behavioral intentions while effort expectancy and social influence affect use behavior~\cite{wei2024using}.

Recent work has taken a deeper look at how users perceive robotaxis beyond acceptance, such as trust~\cite{liu2025understanding,hua2022establish}, perceived safety~\cite{gu2024different,liu2025understanding}, and ethics~\cite{jia2024will}. Liu et al. found that system transparency, perceived ride control, brand reputation, and in-vehicle comfort all shape users' safety perception and trust, which vary by use scenario - nighttime users rely more on ride control and brand reputation, while daytime users prioritize in-vehicle comfort~\cite{liu2025understanding}. Gu et al. found that different Human-Machine Interface (HMI) systems influence the perceived safety of robotaxis passengers, with detailed textual explanations helping maintain stable safety perceptions during incidents~\cite{gu2024different}. Jia et al. found that users regard the trolley dilemma (e.g., protecting passengers or pedestrians) as an ethical issue of high relevance to robotaxis, and prefer robotaxis with an egocentric algorithm that prioritizes passenger safety~\cite{jia2024will}.


Most existing studies examine human factors in robotaxi adoption and use from a macro-level perspective, with limited attention to how micro-level robotaxi design features shape user experience. To address this gap, this work combines interviews and autoethnography to systematically examine real-world users' motivations, experiences, perceptions and expectations of robotaxi use, informing a user-driven design framework for robotaxis.

\subsection{User-friendly Robotaxi Design}

Though most existing work on enhancing robotaxi experience has focused on algorithm development, such as environmental perception~\cite{rajaram2022development,tanaka2025domain}, dispatch optimization~\cite{qi2024multi,li2025optimization}, and passenger identification~\cite{wu2025enhancing}, recent studies have begun to take a more human-centered perspective to investigate robotaxi design~\cite{al2024navigating,ma2025designing,kim2024timelytale,brandebusemeyer2022travelers}. 

One line of work combines interviews with prototyping for the exploratory design of robotaxi for multiple objectives, such as user trust and in-car experience~\cite{hua2022establish,auhspace,ma2025designing}. For example, Hua interviewed 12 participants with a background in the automobile industry and summarized several design components to establish robotaxi trustworthiness, such as feature explainability, decision transparency, and traffic adaptation ability~\cite{hua2022establish}. Auh et al. had semi-structured interviews with 7 participants and conducted an exploratory design to make robotaxi a liminal space transitioning from work to home with meaning, and propose an immersive environment and multi-modal interaction with the ritual
aspect to bring reflective activity after work for finding meaning and releasing emotion~\cite{auhspace}.

Some other researchers adopted survey studies to quantify user acceptance of different design features of robotaxi~\cite{liu2025driving,gu2024promote}. Liu et al. took a scenario-based survey to examine how user acceptance varied across distinct emotional engagement features in robotaxis, and found that scenarios offering minimal emotional engagement got the highest levels of user trust~\cite{liu2025driving}. Gu et al. surveyed 880 respondents and used structural equation modeling to analyze user preferences, revealing the value of computer-based entertainment and inclusive design~\cite{gu2024promote}.

Finally, since large-scale robotaxi deployment has yet to mature, researchers have turned to simulation studies to investigate the robotaxi design space~\cite{brandt2026communicating,meurer2020wizard,lim2021ux,yoo2024anxiety,meister2025external,weschke2021asking}. For example, Brandt et al. examined how different levels of media richness of in-vehicle HMIs (text only, combining text and audio, and further adding a human avatar) influence user experience in robotaxis through virtual reality (VR), uncovering the significant increases in the hedonic quality of user experience and trust with increased media richness~\cite{brandt2026communicating}.
Among simulation approaches, Wizard of Oz (WOZ) studies are widely used to approximate real-world robotaxi experiences and inform interaction design~\cite{meurer2020wizard,lim2021ux,yoo2024anxiety,meister2025external,weschke2021asking}. Meurer et al. conducted a WOZ study with 33 rides from 10 participants where the driver was hidden from the passenger, and provided insights into four design themes, including short-term domestication, the active passenger, passenger experience journey, and dealing with breakdowns across different robotaxi stages~\cite{meurer2020wizard}.
Lim et al. used the WOZ approach to conduct a usability test with nine participants, and found that users expected the built-in screen to display essential information on the trip and the car’s status, the support of voice interactions, and the option to easily escape from the vehicle in case of emergency~\cite{lim2021ux}. Yoo et al. used the WOZ method to identify anxiety-inducing factors in robotaxi use, and proposed HMIs to mitigate anxiety, such as direction guidance, emergency stop, speed control, and AI speaker~\cite{yoo2024anxiety}.

Despite the important groundwork laid by early-stage studies, research conducted in simulated or hypothetical settings is inherently constrained. Leveraging data from multiple cities with early commercial robotaxi deployment, this work presents the first empirical investigation of users’ experiences, perceptions, and challenges in interacting with real-world robotaxis, offering practice-driven insights for the design of future robotaxi systems.


\section{METHOD}
We employed a mixed qualitative approach to address our research questions, integrating semi-structured interviews with robotaxi users and an autoethnographic study to combine participants' reflective accounts with researchers’ situated, experiential observations.

\subsection{Semi-structured Interview}
We adopted semi-structured interviews due to their exploratory, user-centered orientation, which enabled in-depth examination of participants’ perceptions and experiences with robotaxis while allowing flexible follow-up questions.

\subsubsection{Participant recruitment}

The eligibility criterion required participants to have taken a robotaxi ride within the past six months. We chose this time window to balance the currency of robotaxi practices with recall quality and sample size. 


We situated our study in China, one of the most active regions for robotaxi development and deployment~\cite{ponyai,ApolloGo,WeRide}. Given that commercial robotaxi services were deployed only in a limited number of cities at the time of study, we recruited participants through broad social media channels to reach users across multiple urban contexts. In particular, we advertised the study on popular social media platforms in China that were commonly used for sharing and receiving everyday experiences, such as Rednote and WeChat. We also employed a snowball sampling strategy \cite{biernacki1981snowball}. In the recruitment advertisements, we introduced the research topic, interview format, and approximate interview duration, and provided a link and QR code to a screening survey for interested participants. The screening survey collected basic demographic information such as age, gender, and contact information, as well as robotaxi-related background information such as whether participants had taken a robotaxi ride within the past six months, the robotaxi platform(s) used, city of use, and number of rides, which enabled us to recruit participants with diverse demographic and usage backgrounds. Recruitment, interviews, and analysis were conducted on an iterative basis. When we had interviewed around 15 users, we observed that their responses had reached thematic saturation. We then recruited three additional users to confirm this observation and then stopped recruitment. In total, we interviewed 18 users between October and November 2025.



\begin{table}[htbp]
\caption{Participant demographics and robotaxi usage experience.}
\label{demongraphics}
\centering
\begin{tabular}{p{0.7cm}p{0.7cm}p{1.2cm}p{2cm}p{2cm}p{1.5cm}p{1.3cm}p{1.3cm}}
\hline
\textbf{ID} & \textbf{Age} & \textbf{Gender} & \textbf{Education} &
\textbf{City of Use} & \textbf{Service Platform} & \textbf{Ride Count} & \textbf{Last Used} \\
\hline
P1  & 22 & F & Bachelor & Wuhan              & Apollo Go                      & 1     & 04/2025 \\
P2  & 21 & F & High School       & Wuhan              & Apollo Go                      & 3--5  & 05/2025 \\
P3  & 25 & F & Bachelor & Chongqing          & Apollo Go                      & 5     & 05/2025 \\
P4  & 24 & M & Bachelor & Dongguan, Shenzhen & Apollo Go, WeRide, Pony.ai     & 4--5  & 09/2025 \\
P5  & 28 & M & Bachelor & Wuhan              & Apollo Go                      & 20--30& 07/2025 \\
P6  & 24 & F & Bachelor & Chongqing          & Apollo Go                      & 67    & 10/2025 \\
P7  & 25 & M & Bachelor & Wuhan              & Apollo Go                      & 4--5  & 07/2025 \\
P8  & 19 & F & High School       & Wuhan              & Apollo Go                      & 10--20& 10/2025 \\
P9  & 25 & F & Master   & Shanghai           & Apollo Go                      & 2     & 11/2025 \\
P10 & 27 & M & Associate  & Chongqing          & Apollo Go                      & 2     & 09/2025 \\
P11 & 21 & M & High School       & Shenzhen           & Pony.ai                        & 2     & 11/2025 \\
P12 & 28 & M & Bachelor & Guangzhou          & Pony.ai                        & 2--3  & 07/2025 \\
P13 & 23 & M & Bachelor & Wuhan              & Apollo Go                      & 100+  & 11/2025 \\
P14 & 26 & M & Bachelor & Chongqing          & Apollo Go                      & 2     & 07/2025 \\
P15 & 20 & M & High School       & Shenzhen, Wuhan    & Apollo Go                      & 8--9  & 11/2025 \\
P16 & 26 & F & Bachelor & Chongqing          & Apollo Go                      & 1     & 08/2025 \\
P17 & 24 & M & Bachelor & Wuhan              & Apollo Go                      & 2--3  & 04/2025 \\
P18 & 29 & F & Master   & Shenzhen           & Apollo Go                      & 20--30& 10/2025 \\
\hline
\end{tabular}
\end{table}

Table \ref{demongraphics} summarizes participants’ demographics and robotaxi usage experience. Participants' ages ranged from 19 to 29. Of the 18 participants, 8 were women and 10 were men. Most participants held a bachelor’s degree (n = 11), while the remainder had a high school diploma (n = 4), a master’s degree (n = 2), or an associate degree (n = 1). Regarding robotaxi experience, we recruited participants using robotaxis across multiple Chinese cities, including Wuhan (n = 8), Chongqing (n = 5), Shenzhen (n = 4), Guangzhou (n = 1), Shanghai (n = 1), and Dongguan (n = 1). The dominant service platform was Apollo Go (n = 16), and a smaller subset had used Pony.ai (n = 3) and WeRide (n = 1). Self-reported ride counts varied substantially, ranging from fewer than 5 rides to more than 100 rides.

\subsubsection{Interview procedure}

All interviews were conducted via voice calls and lasted an average of 55.54 mins. To ensure that recruited participants had genuinely experienced robotaxi services, prior to the interview, we asked participants to share any ride-related materials they felt comfortable providing (e.g., app screenshots captured during use, and trip records). To protect participants’ privacy, we explicitly encouraged them to redact or withhold any information they considered sensitive. The interview protocol was organized into ten sections. 
\begin{enumerate}
    \item At the beginning of each interview, we explained the research topic, interview procedures, approximate duration, potential benefits and risks, and data use. Participants were given the opportunity to ask questions.
    \item We then began by establishing participants’ general taxi and ride-hailing background, such as their frequency of taxi use and commonly used ride-hailing applications. 
    \item Next, we transitioned to robotaxi-related experiences, focusing on general usage patterns such as the number of robotaxi rides taken and the cities in which these services were used.
    \item We subsequently explored participants’ motivations for using robotaxi services, such as their reasons for choosing robotaxis over other transportation options.
    \item Following this, we examined participants’ perceived benefits and challenges of robotaxi services, such as comparison with traditional taxis and human-driven ride-hailing services. 
    \item We then focused on safety perceptions, asking participants questions such as whether they considered robotaxi services safe, how their safety perceptions compared across different transportation modes, and whether they had experienced moments of concern.
    \item The interview further addressed privacy perceptions, such as whether participants viewed robotaxis as a private space, and how privacy concerns differed from rides involving human drivers. 
    \item We also explored trust, such as whether participants trusted robotaxi systems, and under what circumstances their trust increased or diminished.
    \item To help participants recall experiences in greater detail and articulate desired design improvements, we structured part of the interview around different stages of a robotaxi journey: hailing, pick-up, traveling, and drop-off. For each stage, participants were asked to describe their practices (e.g., how they placed an order and what actions were required during boarding), as well as any concerns, unexpected events, or confusion they encountered. We then invited participants to reflect on potential features or design changes they believed could alleviate these issues.
    \item At the end of the interview, participants were encouraged to share any additional insights or experiences that had not been covered earlier.
\end{enumerate}
The full interview protocol is provided in Appendix \ref{interview protocol}. While we followed this protocol as a guiding structure, the semi-structured nature of the interviews allowed flexibility. The order and emphasis of sections varied across interviews, follow-up questions were used to probe emerging topics in depth, and some questions were omitted when participants had already addressed the relevant information earlier in the interview.

Prior to the formal data collection, we conducted four pilot interviews to assess the clarity of the interview questions and the overall interview duration. Based on feedback from pilot participants, we made minor adjustments to improve interview flow and removed redundant questions that elicited overlapping responses. By the fourth pilot interview, no participants reported confusion or offered further suggestions. As these revisions were minor and primarily aimed at improving clarity, data from the pilot interviews were included in the final analysis, consistent with common qualitative research practice with minimal protocol changes \cite{haney2021s}.


With participants’ consent, all interviews were audio-recorded and subsequently transcribed using Feishu Minutes\cite{feishu}. Upon completion of each interview, participants were compensated for their time with 50 yuan (US\$6.9).\footnote{This amount was set to exceed the local hourly minimum wage in the province of each participant’s city.}

\subsubsection{Interview data analysis}

We employed an inductive thematic analysis \cite{braun2006using}. Two authors independently analyzed half of the transcripts using an open coding approach \cite{corbin2014basics}, separately developing initial codebooks in which codes emerged organically to reflect users’ perceptions and experiences of robotaxis. Examples of initial codes included \textit{limited flexibility in pick-up point} and \textit{slow speed due to over-cautiousness}. The authors then engaged in an iterative process of comparison and discussion, and repeatedly revisited the transcripts to consolidate the two codebooks. Disagreements were resolved through multiple rounds of discussion. We subsequently applied affinity diagramming \cite{muller2014curiosity} to organize conceptually related codes into higher-level themes to produce the final codebook. An example theme was 
\textit{challenge–limited flexibility}. Because the analysis progressed through repeated cycles of coding, discussion, and refinement, codes functioned as an evolving analytic tool rather than a finalized outcome. Consistent with McDonald et al.~\cite{mcdonald2019reliability}, we did not calculate inter-rater reliability (IRR) given the iterative and process-oriented nature of our coding. All quotations were first translated from Chinese into English using Google Translate, after which the first author reviewed and refined the translations to ensure accuracy.


\subsection{Autoethnography}
\subsubsection{Motivation for autoethnography}


First, considerations of generalizability and market structure motivated our use of autoethnography. During our interviews, we found that Apollo Go dominated market adoption, while users of other commercial robotaxi platforms were difficult to recruit. However, our goal was not to conduct a user experience study of a single company, but to examine users’ overall experiences and perceptions of robotaxi systems and to propose a user-driven design framework for robotaxis. Thus, it was necessary for the researcher to engage with multiple commercially available robotaxi systems. As advocated by prior work \cite{cunningham2005autoethnography,o2014gaining,cecchinato2017always}, autoethnography enables first-hand, situated user experience of technologies that are not yet widely accessible or systematically studied, while also supporting more nuanced interpretation of interview findings. 

Second, autoethnography enables systematic and reflexive documentation of experiences across diverse contexts. Many users interacted with robotaxi services only sporadically, which might limit their ability to form comprehensive or well-articulated reflections on these systems. Even when participants had taken multiple rides, they rarely documented or systematically reflected on how their perceptions varied across different scenarios. Autoethnography allows the researcher to record experiences in a detailed and reflexive manner, making experiential nuances and contextual variations more visible. 

Third, autoethnography and interviews play complementary and mutually validating roles. The two datasets serve reciprocal functions: autoethnography contributes depth and reflexive insight, while interview data provides diversity of perspectives and external grounding. Importantly, this combination also helps mitigate the limitations of each method. A key limitation of autoethnography is that the researcher is also the participant, making it difficult to disentangle naturally occurring behavior from research-influenced behavior \cite{fassl2023can}. Interview data, drawn from independent users, helps counterbalance this concern by offering perspectives unaffected by the researcher’s dual role.

\subsubsection{Autoethnographic data collection}

The first and second authors conducted autoethnographic data collection between October 2025 and January 2026 through first-person experiences of robotaxis. Recognizing that perceptions and experiences of robotaxis might be influenced by diverse contexts, and to support a thick and reflexive understanding of these experiences, we attended to a range of ride situations as they naturally emerged in everyday use. Specifically, our experiences unfolded across different travel and road conditions, such as variations in route complexity (e.g., simple straight routes and complex intersections) and traffic density (e.g., low and high). We also tried to encounter temporal variations (e.g., daytime and nighttime) and weather conditions (e.g., clear and rainy). In addition, we paid attention to researchers’ own positional states as riders, such as trip purpose (e.g., commuting and exploratory), familiarity with robotaxis (e.g., the first time and the tenth time), and emotional states (e.g., anxiety, neutrality, and curiosity). We further accounted for social context (e.g., riding alone or with others), spatial positioning within the vehicle (e.g., rear-left seat, rear-right seat, or front passenger seat), and platform used (e.g., Apollo Go, Pony.ai, and WeRide). Importantly, our goal was not to maximize the number of rides or to exhaustively cover all possible scenarios, but rather to cultivate experiential diversity that could enrich reflection, interpretation, and analytic depth.

Data were collected through semi-structured diary entries designed to capture experiential detail and researchers' reflections. Each diary entry was organized into three parts.
\begin{enumerate}
    \item Researchers recorded basic contextual information, including researcher ID; travel and road environment (e.g., complex routes and high traffic density); time, date, and weather (e.g., October 31, 3:00 pm, sunny); researcher state including trip purpose (e.g., commuting), number of prior rides (e.g., the 10th time) and emotional state (e.g., relaxing); social context (e.g., riding alone); spatial positioning within the vehicle (e.g., rear-left seat); the platform used (e.g., Apollo Go); in-cabin conditions (e.g., entertainment features, window status, and temperature settings); and trip characteristics (e.g., waiting time, punctuality, dispatch distance, and travel duration).
    \item Researchers documented immediate reactions and sensations during riding, using voice memos, brief notes, photographs, screenshots, or short video recordings. For example, one voice memo noted, \textit{``Although the robotaxi kept reminding me to buckle up, it started moving before I had fastened my seat belt. Some vehicles will not start until the seat belt is fastened, whereas this one did not.''}
    \item Researchers produced a more complete narrative and reflective account guided by a set of analytic prompts, which is a useful practice in autoethnography for documenting experience data in rich detail \cite{li2024meditating}: (i) a detailed description of what happened during the ride, with attention to moments that felt noticeable, surprising, or emotionally salient; (ii) embodied and emotional experiences, capturing how the ride felt both emotionally and bodily; (iii) interpretations of the system’s behavior, especially moments when actions felt clear, confusing, or unexpected; and (iv) comparisons with prior rides, examining whether and how this experience confirmed, challenged, or shifted prior expectations.
\end{enumerate}

In total, we completed 22 robotaxi rides, with the first author taking 12 rides alone and two researchers taking 10 rides together.



\subsubsection{Autoethnographic data analysis}
Our primary data for analysis consisted of the researchers’ autoethnographic diaries. Additional materials (e.g., screenshots and video recordings) were used mainly to support researchers’ recall of specific situations and experiential details. We also employed an inductive thematic analysis \cite{braun2006using} for autoethnographic data. The first author conducted open coding \cite{corbin2014basics} by reading and coding diary entries to generate initial codes. Because a codebook had already been developed from our prior interview study, the analytic focus was not on building an entirely new codebook, but on supplementing and validating the interview-derived codes. Specifically, supplementation involved identifying codes that were not mentioned by interview participants (e.g., customizing the vehicle’s rooftop light color via the app could signal personal ownership, thereby supplementing users’ sense of agency in Sec. \ref{PA-agency}), as well as elaborating on codes raised in interviews by adding situational nuances that participants had not articulated in depth (e.g., robotaxis providing inadequate user education and support for handling emergency situations in Sec. \ref{PC-Emergency Concern}). Validation focused on examining whether participants’ responses remained valid (e.g., some participants reported that heated seats were unavailable, whereas our autoethnography suggested that this feature had since been introduced on certain vehicles) and accurate (e.g., we identified misunderstandings in how participants interpreted system behaviors, such as assuming that the in-cabin camera is only used for calls and does not record in Sec. \ref{F-privacy}). Invalid codes were removed. For codes reflecting misunderstandings, rather than directly correcting them, we treated them as meaningful perceptions, as they might reflect how robotaxis were perceived by the public. Subsequently, the second author examined these codes and critically reflected on how these experiences had been interpreted.
 Through iterative discussions over multiple meetings, the codes were progressively refined, followed by affinity diagramming \cite{muller2014curiosity} to integrate conceptually related codes 
based on the codebook in the interview study. Data collection ceased when no new themes emerged.

\subsection{Ethical Considerations}

This research was approved by the Institutional Review Board (IRB) of our institution. Informed consent of interviewees was obtained before the official beginning of the interviews. Interview data were securely stored without personal identifiers, and each participant was assigned a pseudonymous label (e.g., ``P06''). Data access was limited exclusively to the research team. Participation was voluntary, and participants were informed of their right to decline any question or to withdraw from the study at any time.

\section{FINDINGS}\label{Finding}
\subsection{General Robotaxi Riding Process}
In this section, we describe the general robotaxi riding flow, synthesized from our autoethnographic experiences and interview data. Rather than documenting platform-specific features exhaustively, we provide a common interactional structure that captures key stages across robotaxi services. The riding process can be broadly divided into four stages: hailing, pick-up, traveling, and drop-off.

The process begins with hailing. Users initiate a ride through a robotaxi platform’s mobile application or mini-program by entering or dragging to a pick-up location and specifying a destination. The interface typically provides an estimated fare before users confirm the request. Once a vehicle responds, users enter a waiting phase, during which the interface displays the vehicle information such as the estimated arrival time, the vehicle’s distance, the approaching route, and vehicle identification information (e.g., a vehicle ID). During this stage, users may also adjust basic in-car settings in advance, such as air conditioning temperature. 

The pick-up stage occurs once the vehicle arrives. Robotaxis usually allow a limited waiting time, within which users must locate the vehicle. Various authentication or unlocking methods are provided to confirm the match between the user and vehicle, such as Bluetooth unlocking, entering the last digits of a phone number, or unlocking directly through the app.

After boarding, users are required to fasten their seatbelts and issue a command to start the ride, after which the journey begins. During the trip, users may engage with the in-vehicle system through the screen or voice interaction, such as viewing the real-time route and remaining travel time, accessing in-car features (e.g., music or seat massage), or contacting customer support when needed.

Finally, during the drop-off stage, users receive notifications as the vehicle approaches the destination. The system prompts users to prepare for exit, alerts them to surrounding traffic, and enables door unlocking for disembarkation.

\subsection{Motivation}
\label{motivation}
\textbf{Low cost} was a key driver. Compared with conventional ride-hailing services, robotaxis were perceived as significantly cheaper. For example, P1 noted, \textit{``If we took a taxi, it might cost around 30 RMB, but taking Apollo Go only costs seven or eight RMB.''} In addition, aggressive promotional strategies further lower the economic barrier to trial. As P2 explained, \textit{``Because they are currently promoting their robotaxi service, they sometimes give us a lot of discount coupons, 30\% or 40\% off, which makes it very cheap.''}

\textbf{Recommendations from acquaintances} played an important role in encouraging first-time use. Trust in peers reduced the uncertainty associated with this novel technology. As P1 described, \textit{``A classmate asked me to hang out and told me about robotaxis. We thought it was something new, so we decided to try it.''}

\textbf{Curiosity} toward the novelty of robotaxis motivated initial adoption. P13 stated, \textit{``The first time was purely out of curiosity.''} Moreover, this curiosity was often cultivated through its exposure on social media. For instance, P18 explained, \textit{``I saw robotaxis online, and I also watched some videos people posted introducing them. They felt really high-tech. I'm curious, so I wanted to try it out.''}



\subsection{Perceived Advantages}

\subsubsection{A sense of agency.}\label{PA-agency}

First, agency was manifested through operational control over the in-car space. Users were able to customize the in-car settings according to their own preferences, such as adjusting the temperature, choosing the music they want to listen to, or opening and closing the windows. In this sense, the vehicle became a space where users could ``do whatever they want''. As P12 explained, the vehicle supports voice commands that allow passengers to directly control various functions, \textit{``I can ask it to turn the volume down, and when I feel hot, I can say I want the temperature lowered.''} The advantages of this form of agency became particularly salient when contrasted with rides that involve a human driver. In conventional ride-hailing services, passengers often lacked autonomy and must communicate their requests to the driver and obtain consent, and such requests were not always accommodated. As P12 noted, \textit{``I’ve encountered some drivers who have habits that negatively affect my riding experience, like playing the music too loudly, choosing music I don’t like, or setting the air conditioning in a way that makes me uncomfortable. But since I’m an introverted person, I often feel awkward about making these requests or suggestions.''} Similarly, P1 mentioned, \textit{``Some cars smell really bad. And you can’t even open the window. If you do, the driver will complain and say the air conditioning is on.''} 

Second, agency was experienced as a form of psychological freedom and ownership over the vehicle. Participants described feeling mentally more relaxed and less restrained in robotaxis, perceiving the car as ``their own'' rather than someone else’s property. As P16 explained, \textit{``I feel more free to speak my mind in a driverless car. With Didi\footnote{Didi is a popular Chinese mobility technology company, offering app-based transportation and related services such as ride-hailing. \url{https://en.wikipedia.org/wiki/DiDi}.}, it always feels like this isn’t my private car and it’s the driver’s car. But with Apollo Go, it feels like this is the car I ordered, and I can say whatever I want inside.''} This sense of ownership was further reinforced through personalization features. In our autoethnographic observations, we found that robotaxi users could customize the color of the vehicle’s rooftop light via the app. When waiting for the car, the approaching vehicle displayed the user’s chosen color, creating a sense of exclusivity and personal connection. P12 also highlighted this experience, \textit{``I think being able to customize things like colors makes you feel more comfortable. It’s like every ride has a color that belongs to you and waits for you. It gives you a sense of ritual.''} Additionally, robotaxis reduced the social and communicative burden often associated with human drivers. As P11 explained, \textit{``In terms of freedom, it’s better. Some drivers are overly talkative or proactive, and out of politeness, you may feel uncomfortable refusing their enthusiasm. With robotaxis, you can choose to talk to the AI if you want, or not talk at all.''} Similarly, P04 emphasized the value of a personal space, \textit{``Robotaxis don’t require human interaction, and for someone like me who is a bit introverted and prefers minimal social contact, they offer a completely private environment. Inside, I can do whatever I want without interacting with anyone, which makes me feel much freer.''}




\subsubsection{Behavioral consistency and experiential standardization}
This consistency manifested, first, in the robotaxi’s driving behavior. Participants emphasized that robotaxis demonstrated sustained attentiveness while driving. As P11 noted, \textit{``One good thing is that you don’t have to worry about the driver. Sometimes, human drivers might use their phones and get distracted. With robotaxis, you don’t have to worry about that.''} Beyond attentiveness, robotaxis were also perceived as emotionally neutral and rule-abiding. P12 contrasted robotaxis with human drivers who might display aggression or emotional reactions in traffic, \textit{``Some drivers can be irritable, and when someone cuts in, they may start arguing or competing with other drivers. Robotaxis will follow the rules very strictly. For passengers, it feels more rational because it doesn’t show emotions.''} Moreover, robotaxis did not introduce ad hoc or last-minute changes. As P2 explained, \textit{``One advantage of Apollo Go is that the pick-up point is fixed. You just wait there. With Didi, drivers sometimes think the traffic isn’t good and change the pick-up location, asking you to walk a bit further and calling to let you know.''}


In addition to driving behavior, participants described robotaxis as offering a standardized riding experience. P14 observed that with conventional ride-hailing platforms, experiences could vary widely depending on the driver, especially when discounted rides were involved, \textit{``Sometimes you may get a special deal, but the driver won’t even turn on the air conditioning. The riding experience in robotaxis is much more standardized and regulated because it doesn’t vary dramatically from passenger to passenger.''} This standardization was further reflected in the in-vehicle environment. Similarly, P07 appreciated that \textit{``There’s no driver smoking in the car, which I really dislike, and none of that swearing or rude behavior you sometimes get with drivers.''} Finally, participants associated robotaxis with a consistently polite climate compared with conventional ride-hailing services. P16 pointed out that \textit{``Didi drivers usually don’t offer any reminders, such as welcoming you when you get in, reminding you to fasten your seatbelt, or telling you to be careful when getting out.''} 


\subsubsection{Transparency.} \label{PA-transparency} Another advantage was transparency with respect to both route and pricing information. In terms of routing, participants emphasized the reliability of robotaxis. P18 highlighted that \textit{``Robotaxis won’t deliberately take detours,''} while P2 noted, \textit{``Compared with traditional taxis, the driving route of robotaxis is very explicit. I can see exactly which route the vehicle is following.''} With regard to pricing, participants perceived robotaxi fares as standardized and free from hidden charges. Unlike traditional taxis or ride-hailing services, where drivers may request additional fees under certain conditions, robotaxis were seen as adhering to predefined pricing rules. As P06 explained, \textit{``Some Didi drivers may ask you to add a return-trip fee, for example, five yuan, or even ten yuan during holidays. They feel the distance is too long and that they won’t be able to pick up passengers on the way back, so they charge an extra return fee. This used to be more common with taxis, but now some Didi drivers do it as well. With Apollo Go, this problem doesn’t exist. It never asks me to pay an additional return fee.''} 



\subsection{Challenges}
\subsubsection{Limited Flexibility}\label{PC-Flexibility} 
\textbf{Driving behavior.} First, participants perceived the robotaxi’s driving style as overly cautious and conservative, which often led to efficiency loss. For example, P2 described that \textit{``It keeps letting others go, and only moves when it feels it’s absolutely safe. I remember I saw it stop there for more than 30 minutes.''} This conservative driving style also made the robotaxi vulnerable to being cut off by other vehicles. As P12 explained, \textit{``It doesn’t dare to compete for lanes or cut in, and other cars dare to cut in front of it. When that happens, it drives very slowly. If you’re in a hurry, you definitely wouldn’t choose it.''} Participants further noted that excessive caution and yielding behavior often resulted in frequent hard braking, which would negatively affect passengers' ride comfort. P1 mentioned, \textit{``In Wuhan, drivers won’t necessarily yield to you. Since it follows traffic rules more strictly, it ends up braking hard all the time when other cars change lanes or overtake.''} 

Beyond specific driving maneuvers, some participants reflected more broadly on the efficiency costs of strict rule compliance and questioned when rules should be followed rigidly versus flexibly. P10 compared robotaxis with human-driven taxis and highlighted the trade-off between legality and convenience, \textit{``It completely follows traffic rules. In some places, you obviously can’t cross a solid line or make a U-turn, but a normal taxi would just turn there and come back quickly. The robotaxi has to drive ahead, maybe another kilometer, make a legal U-turn, and then come back. Following the rules is not wrong, but what’s sacrificed is our waiting time.''} P10 further argued for a balanced approach between rigid rule compliance and situational adaptability, \textit{``Ideally, there should be a middle ground between rules and flexibility. In some situations, it strictly follows traffic rules, and in some other situations, it can slightly bend them. There needs to be a clear boundary. Of course, this should only happen when passenger safety is guaranteed.''}

Second, limited flexibility in driving behavior was also reflected in the robotaxi’s difficulty in understanding the intentions of pedestrians and other vehicles. P06 noted that \textit{``When you’re driving yourself, you can judge how a pedestrian is going to move, or whether they want to cross. You can kind of feel it. But the robotaxi can’t judge that. It can only see that there’s someone or something there, but it can’t make accurate judgments like a human.''}

\textbf{Parking behavior.} The inflexibility of parking behavior was a persistent challenge. One major issue was that pick-up and drop-off points were fixed. Many users noted that robotaxi services relied on predefined stations and passengers must walk to these designated locations to get on, which differed fundamentally from ride-hailing services or taxis with drivers, where vehicles could typically stop almost anywhere. Several users described robotaxis as having ``invisible bus stops'', which substantially undermined convenience. As P09 explained, \textit{``That means when you want to take a taxi, you first have to walk ten or even twenty minutes to that fixed point. Originally, taking a taxi was supposed to be convenient, but with this robotaxi, you have to walk a long distance just to reach the pick-up point. I think this is one of its major inconveniences at the moment.''} P05 also described the absurd situation of passengers having to chase the vehicle, 
\begin{quote}
    \textit{``If you are not waiting at that exact point, even if you see the car on the road, it will just pass right by you. You have to go to that precise waiting spot to get on. It doesn’t matter whether it detects your phone’s location or not. It just zooms past you, and you end up running to catch it and get on.''}
\end{quote}
Furthermore, their specific positioning was not always accurate. P14 noted, \textit{``When it planned to pick me up, it had driven about 50 meters ahead of the pick-up point. When dropping me off, it also stopped 30 to 50 meters beyond the drop-off point. I didn’t expect the deviation to be that large.''} We observed similar issues in our autoethnography, and inaccurate localization made the researcher feel uneasy during the ride.


In such situations, some platforms claimed to allow users to select their own pick-up and drop-off points anywhere, but these points often failed to account for surrounding conditions. P04 described the inconvenience caused by poorly chosen stopping locations.

\begin{quote}
    \textit{``It dropped me at a place where there were guardrails for 200 meters, so I couldn’t even get through. I had to walk along the road, with cars passing right next to me. They advertise that you can get on and off at any point, but I feel like this feature sacrifices user experience just to force a `stationless' function. Although it's me who chose that point in the app, I couldn’t tell from the interface that there were guardrails there. Besides, you can’t tell the robotaxi, `Could you go a bit further forward?' Once it reaches the point, it just makes you get off.''}
\end{quote}

In our autoethnographic, we found that Pony.ai allowed users to slightly adjust the vehicle’s stopping position on the mobile interface at pick-up by dragging the target parking spot. However, we noticed that we had little sense of how the on-screen distance corresponded to real-world distance and must rely on intuition when dragging. Moreover, the adjustment was limited to forward and backward movement, with no option to shift laterally. While this feature partially mitigated the inconvenience of poorly positioned pick-up points, it introduced new concerns. For example, temporarily granting users a degree of control over the vehicle raised questions about responsibility.

In addition, users highlighted the rigid waiting-time policy after vehicle arrival. Robotaxis typically waited only a short period before departing if passengers failed to board on time. P03 described this trade-off between strict waiting-time policy and real-world traffic conditions, \textit{``There are pros and cons. In some places, stopping for too long can block major traffic arteries and cause congestion. But the downside is that sometimes the car leaves when you are just one or two seconds late.''}

\textbf{Service.}
Robotaxi services lacked the flexible assistance service that human drivers could provide. A commonly raised issue concerned handling luggage. Some users expected drivers to help load luggage into the trunk, but this was not something robotaxis could offer. As P03 noted, \textit{``If you are a woman, a driver can help you lift the luggage, because sometimes it’s quite heavy and needs to be put into the trunk. But Apollo Go can’t do that.''}

\textbf{Trip customization.} Robotaxis failed to accommodate many users’ customized needs during the trip. Several users noted that robotaxis did not support temporary stops followed by trip continuation. As P05 explained, \textit{``If I want to stop the car to buy a bottle of water or pick up a friend, you can usually just say a word to a human driver. In contrast, such ad hoc adjustments are not possible with robotaxis.''} Users were also unable to choose among alternative routes. As P11 noted, \textit{``Right now, it can only follow the platform’s predefined route. There’s only one route, and there’s no way to choose another one.''} In addition, robotaxis did not allow users to select different driving modes based on situational needs. Regardless of time or contextual requirements, passengers must adapt to a single mode of operation. P10 expressed a desire for differentiated driving modes, \textit{``I think I’d like to have options. For example, you could drive faster without worrying too much about comfort if I’m in a hurry. Or if there are elderly people or a pregnant passenger, you could drive more smoothly.''}

\subsubsection{Limited Transparency.}\label{PC-Transparency} Although some participants identified transparency as a perceived advantage of robotaxis over traditional taxis or ride-hailing services in Sec. \ref{PA-transparency}, users nonetheless reported uncertainty and discomfort arising from insufficient access to timely, operational, financial, and vehicle-related information. First, temporal transparency remained limited. After placing a request but before the ride was accepted, users were not informed how long they might need to wait for a vehicle to respond. As P04 complained, \textit{``When there’s temporarily no car accepting the order, it doesn’t show you a specific time at all. You don’t know how long you’ll be staying on that screen. It’s unknown, completely unknown.''} Even after an order was accepted and the app displayed an estimated arrival time, robotaxis frequently exhibited discrepancies between estimated and actual arrival times. P06 described this issue and her subsequent adaptation, \textit{``It might tell you five minutes, but it actually takes ten minutes to arrive. So now, when I book Apollo Go, I always do it a few minutes earlier.''} Our autoethnographic observations corroborated this experience. This recurring mismatch deepened our doubts about the accuracy of the estimation system and raised questions about whether optimistic estimates functioned as a marketing strategy to reduce perceived waiting time.

Second, users pointed to a lack of operational transparency regarding vehicle availability. Passengers could not see how far the nearest available vehicle was or how long it would take to reach them, which was the information they considered crucial for deciding whether to choose a robotaxi. As P06 suggested, \textit{``It could show you the nearest available car and how many minutes it would take to get to you.''}

Third, price transparency was insufficient across platforms. Not all robotaxi platforms provided clear or detailed pricing rules. In our review of Pony.ai’s app and through follow-up communication with customer service, we found that pricing explanations remained vague. Estimated fares were described as being calculated based on real-time traffic conditions, estimated distance, and time, but no breakdown was provided (e.g., cost per kilometer or per minute). Moreover, our ride experiences revealed substantial price variation over time, with initial rides being extremely inexpensive, sometimes costing only a few yuan much lower than conventional ride-hailing services, while later fares approached nearly twice the price of conventional ride-hailing services. P06 explicitly described this as a form of price discrimination against regular users, \textit{``I do think there’s something like `killing loyal customers.' The discounts just aren’t as good as they used to be.''}

Finally, users reported limited transparency regarding vehicle condition and system status. There was a lack of explicit information about the vehicle’s operational health and corresponding explanations. As P09 recounted, \textit{``I don’t know its health condition, such as whether the battery is fully charged and what the tire pressure is like, and none of that is shown. The second time I took it, the dashboard in front of the steering wheel showed `sensor lost,' but I didn’t really understand what that meant. My first reaction was to wonder what was going on. If I were more nervous, it could actually be quite scary.''}

\subsubsection{Management Difficulty}

Robotaxi platforms also faced a range of management challenges in day-to-day operations. One key challenge concerned vehicle cleanliness and hygiene. Although many users regarded hygiene as a benefit of robotaxis and often assumed them to be cleaner than human-driven vehicles (e.g., P08, P09, and P17), our auto-ethnographic observations suggested that this benefit also met difficulties. We encountered several robotaxis with leftover trash and unpleasant odors. While some platforms automatically ventilated the vehicle by opening windows after passengers disembarked, waste management remained unresolved.

Another management issue involved lost-and-found handling. We observed that Apollo Go was capable of automatically detecting items left behind in the vehicle and providing in-car voice reminders to alert passengers. However, this mechanism only functioned if passengers were still present. Once a passenger had already walked away, in-vehicle voice prompts could not effectively address the situation.

Moreover, long waiting times also reflected the difficulty of determining appropriate fleet size and dispatch strategies. Many users expressed frustration with prolonged pickup times compared with traditional ride-hailing services. For example, P10 noted that \textit{``The pickup time was quite long, and the assigned vehicle was sometimes very far away, four or five kilometers, so it could take around ten minutes to arrive, which was quite long.''} 



\subsubsection{Robustness Concern}\label{PC-Robustness}
Users expressed concerns about the robustness of robotaxis, especially in unusual or edge-case conditions. For example, P15 described an incident in which the vehicle came to a complete stop when encountering an unexpected obstacle, \textit{``Once there was something right in front of the car, some kind of bag. It might have been blown up by the wind, so it looked quite large. The car immediately turned on the hazard lights and stopped in the middle of the road.''} Another type of edge case involves encounters between two robotaxis. P15 recounted, 
\begin{quote}
    \textit{``I once reached a turn where there was another robotaxi coming from the opposite side. The two cars got stuck there, and neither moved. It lasted for probably ten or twenty minutes. I was really confused about why it suddenly stopped. The screen just kept showing `there is a car ahead,' meaning you can’t pass and have to wait for the car in front to move before turning around. The other car probably received the same prompt, so both vehicles just stayed there without moving.''}
\end{quote}

Moreover, users worried that extreme weather conditions and specific times of day may undermine system robustness. P17 noted that \textit{``Another issue is handling special tasks that the robotaxis haven’t been trained on. For example, if heavy fog combined with complex road conditions, the system may not have learned how to handle that. What works on a clear day may fail in fog. Many algorithms could become invalid, and you’d need to activate contingency plans. For instance, overtaking might be fine in sunny conditions, but can it still make correct judgments in dense fog? That’s very important.''} Similarly, P06 expressed concern about nighttime operation, \textit{``If it’s too dark, I worry that it might not be able to recognize things properly.''} 


In addition, users also raised concerns about technical vulnerabilities, such as network reliability and cybersecurity. P17 mentioned worries about network coverage, \textit{``In some places, network coverage may not be guaranteed, which is definitely a concern.''} The same participant also expressed fears about hacking, \textit{``I’m also afraid these systems could be manipulated by spies or hackers who modify the program to cause harm or even murder. If hackers break into the system, this could potentially happen on a large scale.''}

\subsubsection{Emergency Concern}
\label{PC-Emergency Concern}
Robotaxi systems provided inadequate user education and support for handling emergency situations. During one of our researchers’ ride experiences, an unexpected incident occurred. The first author noticed that the vehicle was driving abnormally slowly, well beyond a reasonable range, which triggered panic and a sense of helplessness. The instinctive reaction was to stop the vehicle and get out. However, there was no clearly visible emergency button in the cabin. In that moment of fear, the researcher experienced cognitive overload and was unsure how to act. Eventually, a phone icon on the screen was discovered. After connecting to customer service and explaining the situation, a remote safety driver took over and brought the vehicle to a stop. 

Motivated by this experience, researchers conducted a focused observation of emergency and safety-related design. We identified several potential safety risks. First, although robotaxis typically presented a brief onboarding or safety video at the start of a ride, this information was not shown consistently on every trip. Second, while vehicles were equipped with emergency hammers, there was no clear, concrete explanation of how or when to use them. Although the placement of the safety hammer indicated that usage instructions were provided in the safety guidelines, neither the safety video nor the safety card offered a concrete explanation of how to use it. Third, critical safety information was poorly visualized. The text used in safety videos was small and stylistically ``cute,'' which failed to attract attention and was difficult to read even when deliberately sought out. Fourth, safety instructions were inconsistent across interfaces. For example, the safety video instructed users to press an emergency door-release button in urgent situations, whereas the physical signage on the door stated that the door could be opened by pulling the handle twice. Fifth, some safety-related controls were insufficiently signposted. For instance, in Pony.ai vehicles, an SOS button was located above the front passenger seat in a visually inconspicuous area, yet we were unable to find any in-vehicle explanation indicating its existence or function. Finally, in the researcher's experience mentioned above, when contacting customer service and the remote safety driver, the researcher as a passenger was not provided with an explanation of the incident nor any useful follow-up feedback.

Participants shared these concerns. P09 specifically highlighted the low visibility of emergency controls and the lack of user education around emergency procedures, \textit{``If there’s an emergency, I think the system must have something, but I didn’t see it. I couldn’t find anything like an SOS or emergency help button in a very obvious place. If there were one, I think that would be much better. Right now, when you get in the car, you usually just enter your phone number, tap `start trip,' and the car begins moving. There’s no introduction at all about what to do in an emergency.''}

\subsection{Perceived Privacy, Safety, Ethics and Trust}

\subsubsection{Privacy perception}\label{F-privacy}

Participants typically expressed indifference toward privacy issues rather than active resistance. The attitude was grounded in the belief that ``There is no privacy anymore.'' P04 pointed out that \textit{``Surveillance is ubiquitous in shopping malls and mobile applications, often without explicit or meaningful notice.''} Against this backdrop, robotaxi monitoring felt unexceptional. Some users argued that they would simply avoid engaging in private or sensitive behaviors inside a robotaxi. P05 said, \textit{``I’m just taking a ride, that’s all. I don’t mind, because knowing the car records audio and video, I wouldn’t express anything private in the first place. I don’t care about the monitoring or audio recording.''} Some participants even misunderstood the in-cabin cameras, stating, \textit{``That camera is for calls, so it doesn’t record anything.''} (P07).

Participants’ privacy indifference was further reinforced by a perceived low individual data value, especially under large-scale data collection. As P04 noted, \textit{``Unlike a human driver who might directly hear and remember our personal conversations, robotaxis collect massive volumes of data, e.g., tens of terabytes per day, where no one cares about what a particular passenger did at a specific moment.''} Some users further expressed that \textit{``Our lives are too boring to be worth monitoring.''} (P06), or \textit{``I don't even understand why the system would need to collect such data.''} (P04). A potential underlying reason for privacy indifference was that users felt they had little control over their privacy. P05 said, \textit{``Even if I'm uncomfortable with data collection, I cannot realistically opt out, since in-cabin audio and video recording is here.''} 


Importantly, when users expressed privacy concerns, these were directed primarily at the risk of data leakage rather than at data collection itself. P09 stated that \textit{``Collection is acceptable as long as the data is not disclosed or misused.''} P07 argued that \textit{``In-cabin data should ideally not be stored at all, and that if storage is unavoidable, strict guarantees against leakage are essential.''} Underlying these concerns was a broader lack of data transparency. Users were often uncertain about whether recording access was real-time or only triggered by incidents, whether user authorization was required, and under what conditions data could be retrieved. P10 asked, \textit{``Can people at the dispatch center directly see what’s happening in every car whenever they want, or is the in-cabin monitoring only accessed after something is triggered, like a problem or an incident? For example, if I lose something, can they review the recording from that specific trip? Or is it the case that every car on the road can be remotely accessed at any time, with the dispatch center watching the interior in real time without us knowing?''} Users therefore expressed a strong expectation for explicit and upfront explanations of privacy practices, emphasizing that clear disclosure would allow them to make an informed and voluntary choice about whether to accept robotaxi monitoring.

At the same time, users were willing to accept data collection when it was framed as functionally necessary and socially beneficial. Examples included monitoring uncivil behaviors (e.g., smoking or eating in the vehicle (P04)), locating lost items (P11), ensuring passenger safety (P08), and supporting system maintenance and continuous improvement (P06).

Participants also distinguished between privacy exposure to human drivers and machine-based monitoring in robotaxis. Although robotaxis were known to continuously record audio and video, this form of surveillance was perceived as less intrusive. Machine-based monitoring was described as implicit and low-profile by some users. P04 shared that \textit{``The cameras are definitely there, and they are certainly recording both video and audio, but it doesn’t feel as intrusive as having a person sitting right in front of you. It’s not that your privacy is truly protected. After all, this is still a public space where you wouldn’t do anything very private anyway, but the sense of intrusion is lower. It feels less like being probed for information and less awkward, because the recording is discreet rather than overt.''}

\subsubsection{Safety perception.}\label{F-Safety} Regarding safety perceptions of robotaxis, users held divergent views. Some users expressed significant concerns about safety. First, concerns stemmed from doubts about robotaxis’ driving competence and decision-making ability, particularly under extreme conditions. As P09 explained, \textit{``Human drivers rely on the brain’s complexity to make rapid and accurate decisions in extreme situations. I don't know whether robotaxis could truly match human reflexes, intuition, and subconscious reactions, which often occur before conscious reasoning. Many emergencies are inherently unpredictable and cannot be fully computed. Moreover, human cognition is more complex and adaptable than commonly assumed, while robotaxis are ultimately constrained by human-written code and therefore cannot be perfect.''}

Second, users reported a diminished sense of personal control in potentially dangerous situations. P09 described a fear associated with the absence of a human operator, \textit{``I will have a sense of helplessness arising from the inability to intervene if something goes wrong.''} This perceived lack of control amplifies anxiety when riding in a fully autonomous vehicle.

Third, some users’ safety concerns were rooted in uncertainty about what and who was behind the robotaxi system, particularly doubting whether system designers possessed sufficient real-world driving experience. As P07 questioned, \textit{``Experienced human drivers with decades of driving history often inspire trust, whereas robotaxis are developed by programmers who may not even hold a driver’s license. I don't know whether traffic rules and real-world driving nuances are encoded by professional drivers or by AI engineers unfamiliar with practical road conditions.''}  This lack of transparency regarding the expertise of system designers contributed to fear.


In contrast, other users perceived robotaxis as safer than human-driven vehicles for different reasons. First, although Sec. \ref{PC-Flexibility} notes that robotaxis may limit flexibility, their law-abiding and cautious driving style could enhance perceived safety. As P08 explained, \textit{``Slower driving is often preferable to aggressive behavior. I had experiences with human ride-hailing drivers who frequently sped, overtook other vehicles, and rushed through yellow lights, especially late at night. In comparison, robotaxis’ conservative driving felt more reassuring, even if it meant longer travel times.''}

In addition, robotaxis eliminated the risk of interpersonal conflict or harm involving drivers, which was particularly salient for female passengers. P03 shared past experiences of disputes with taxi drivers late at night, noting that \textit{``Drivers’ fatigue or irritability sometimes escalated into loud arguments. Although drivers may not perceive their behavior as threatening, such confrontations can be intimidating for women.''} In this context, the absence of a human driver made robotaxis feel substantially safer.



\subsubsection{Ethics perception}\label{F-Ethics}
\textbf{Accountability in an accident.} A primary ethical concern was accountability when an accident occurred. Users expressed uncertainty about who should bear responsibility: the passenger, the autonomous system, or the platform. For example, P15 questioned \textit{``If you take this car and it accidentally hits someone, whose responsibility is it, the passenger who ordered the ride, or the system itself? I feel that it might end up being my responsibility. After all, the autonomous car follows my instructions, there is no human driver, and the accident would not have happened if I had not chosen to take this ride.''} Relatedly, some users felt an increased moral responsibility to remain attentive to road conditions when riding in a robotaxi. P18 said, \textit{``With a robotaxi, during the autonomous driving process, I feel that I personally would be more proactive in paying attention to the road conditions. But when there is a human driver, as a passenger, I think we tend to hand this responsibility over to the driver.''}

Moreover, accountability concerns became particularly salient in relation to customization features. As P18 mentioned, \textit{``I think in the future there may be personalized settings, such as vehicle speed. But for speed, I feel that not customizing it might actually be fine. Once you start defining these settings, if an accident really happens, responsibility becomes hard to determine. If you don’t participate at all and something goes wrong, then it’s probably the system’s responsibility. But if you increase the speed and an accident happens because the vehicle doesn’t react in time, then that responsibility is entirely yours. So I feel that not customizing it might actually be relatively safer.''} Users strongly expressed expectations for clear regulatory frameworks. As P18 noted further, \textit{``If an accident happens, how responsibility should be allocated in the most reasonable way is very important. I hope that the relevant authorities can issue more detailed regulations, so that we know how to protect our rights.''}





\textbf{Responsibility in feedback.} Another ethical issue concerned responsibility in feedback attribution. Some users questioned the fairness of providing a single, overall rating for a robotaxi experience. P17 argued that \textit{``The car is basically standardized, and there isn’t much difference from one car to another. So what exactly are we rating? We could rate things like task handling or the data-processing model. Behind each of these models, there is someone who wrote the code, so giving a low rating is essentially evaluating the code they wrote.''} This reflected an emerging ethical expectation for granular responsibility allocation within complex sociotechnical systems.





\textbf{Delegating life-and-death control.} Users also raised fundamental ethical concerns about ceding control over human life to machines. P17 described, \textit{``It feels like I’m completely handing the fate of a living being over to something that has no life, and that makes me feel uneasy.''} This concern intensified in multi-passenger scenarios. The same participant noted that \textit{``The more people there are, the less likely I am to choose robotaxis. I wouldn’t even dare to entrust one person’s life to it, let alone several people’s. And even if one person agrees, what if the others don’t? If even one person disagrees, then this car really shouldn’t be used.''}


\textbf{Overlearning.} Some users worried that robotaxis may overlearn undesirable driving behaviors, such as aggressive merging or cutting in. P04 reported, \textit{``I expect it to be relatively conservative and well-behaved. So when it suddenly cuts in, I am really shocked. That lane-cutting behavior makes the experience quite unpleasant. It feels like something goes wrong, and it suddenly decides to cut in.''} 

\textbf{Absence of human-like social norms.} Finally, users emphasized that robotaxis failed to adhere to human social driving norms. Beyond rule compliance, they expected robotaxis to demonstrate an understanding of implicit, culturally shared driving etiquette. For example, P17 pointed out that \textit{``Many human drivers follow unspoken conventions that AI systems often fail to recognize. These include flashing headlights to signal thanks, briefly honking to acknowledge another driver after merging, or using short horn signals on S-shaped or blind curves to warn oncoming traffic.''} While such practices were widely understood among human drivers, robotaxis did not perform these social signals, which could make their behavior unnatural or socially disconnected. P10 further highlighted differences in anticipatory driving behavior, \textit{``Human drivers often infer others’ intentions from subtle cues, such as brake lights turning off, and prepare to act in advance, allowing for smoother and more synchronized movement. In contrast, robotaxis tend to wait until the leading vehicle has fully started moving, then re-perceive and re-calculate the environment before acting. Another vehicle may cut in during the hesitation, preventing the robotaxi from moving forward at all.''} Users, therefore, expected robotaxis to develop more human-like anticipation and responsiveness based on fine-grained traffic cues.


\subsubsection{Trust perception}
\label{TP}
Users' trust in robotaxis appeared to be associated with several interrelated factors.

\textbf{Personal riding experience.} Positive and repeated usage experiences contributed to the gradual development of trust. Initial concerns tended to diminish as users gained more experience without encountering negative issues. As P13 noted, \textit{``When I first started using it, I did worry about things like whether it might suddenly run out of power or lose signal. But after using it more often, I realize that I haven’t really encountered these kinds of problems, so I gradually feel more at ease.''}



\textbf{Media exposure.} The absence of negative news, particularly reports of accidents, fostered a baseline level of trust. For example, P06 mentioned, \textit{``I hadn’t heard about any negative incidents, so when I took it for the first time, I approached it with a fairly trusting attitude.''} 


\textbf{Environmental conditions.} Trust was context-dependent and tended to decrease under complex road conditions or adverse weather. P15 expressed that \textit{``When traffic is very congested, I wouldn't choose it because I'm afraid it might get stuck. Likewise, if there are a lot of pedestrians on the road, I wouldn’t take it either.''} Weather was also highlighted as a key factor. P03 mentioned that \textit{``I definitely trust it more when the weather is sunny, because rainy conditions would certainly affect its visibility. Its cameras and sensors would all be impacted.''} Some participants pointed out that robotaxi services were suspended under unfavorable weather conditions. While such withdrawal might be intended as a safety measure, it could also be interpreted by users as a form of avoidance. As P04 explained, \textit{``From what I know, in extreme weather they suspend all operations. When they encounter more challenging environments, they choose to avoid them and just withdraw completely. They don’t try to explore the limits of the algorithm. They simply operate in their comfort zone, and once they move beyond it, they give up.''} From this perspective, suspending operations in difficult environments signaled a reluctance to explore or extend the system’s operational limits, which might in turn reduce users’ trust.

\textbf{Government regulation and institutional oversight.} Users’ trust in robotaxis was partially derived from their trust in governmental regulation and state oversight. Several participants believed that technologies permitted to operate publicly must meet certain safety standards. P10 emphasized that \textit{``If they are allowed to operate on the market, I believe they must have reached a certain level of capability''}.




\textbf{Deployment strategies.} A phased rollout, rather than an immediate, large-scale deployment, was seen as reassuring. Participants expressed higher trust in robotaxis that underwent step-by-step testing, starting with limited regions and safety operators before full public access. P11 described, \textit{``Testing is opened up gradually rather than all at once. It’s rolled out city by city. Because of that, I feel fairly trusting. For example, in some areas, there are safety operators accompanying the tests first, and only after some time is it opened to the public. This kind of orderly regulation makes me feel reassured.''} Similarly, P03 noted that \textit{``Restricting operations to well-tested routes and avoiding highly complex roads contribute to a sense of safety and trust.''}



\textbf{Technical knowledge.} Trust levels also varied based on users’ knowledge of robotaxi technology. Participants with professional or technical backgrounds demonstrated higher confidence, as they understood the system’s capability boundaries. P04 explicitly stated, \textit{``I trust it very much. Because I more or less understand the current capability boundaries of autonomous driving algorithms, I know that in certain road conditions, letting it operate is definitely not a problem. The probability of an accident is actually quite low.''}

\textbf{Safety-related design features.} Trust was further reinforced by observable safety-oriented design choices, such as conservative speed control and strict compliance with traffic rules. P18 mentioned \textit{``Overall, I think its speed is controlled very appropriately, and it seems to follow traffic rules closely. It never overtakes and always yields to pedestrians.''} In addition, interface features helped users perceive that the vehicle could ``see'' and interpret its environment. For example, P06 pointed out that \textit{``On its map, if you zoom in, you can see roadwork indicators. It shows that the road is under construction, with a red construction icon displayed on the map.''}
Finally, enforced safety procedures conveyed a sense of rigor and control, further strengthening user trust in the overall system. P09 shared that \textit{``The overall process feels very safe and quite strict. When you get in, you’re required to fasten your seatbelt. If you don’t buckle up, the vehicle simply won’t start and keeps showing a reminder. When you get out, it reminds you to take your belongings, prepare in advance, and also alerts you to vehicles approaching from behind.''}

\subsection{Expected Designs}

\subsubsection{Control and trust expectations.}
\label{ED-Control and trust expectations}
\textbf{Explainability.} Many users expected robotaxi systems to provide explainable feedback that helped them understand what was happening and why. Although some platforms had incorporated very basic forms of explanation, these remained minimal and repetitive. For example, as P06 noted, \textit{``The system sometimes says something like `the traffic situation is a bit complex', but it seems to always be the same sentence.''} 

Users expected explanations that were more detailed and more closely tied to the current situation, rather than a single and mechanical message. Their primary expectation was for explanations of the vehicle’s driving behaviors. As P13 noted, \textit{``If it suddenly stops, it should tell me why. For example, if it detected an accident ahead, it could explain that and then ask whether I want to change the route.''} Similarly, P08 further explained, \textit{``I would want it to explain, because being alone in a closed space can be a bit scary. I’d hope it could tell me what’s happening ahead, why it’s taking this action, maybe ask me to wait a moment, or tell me how many minutes until we can move again. That would give me a sense of control and make me feel better.''} 



Moreover, users’ expectations regarding when they need explainability from robotaxis appeared to differ. Most participants emphasized the need for explanations in unusual situations, such as accidents, emergency braking or unexplained technical breakdowns. P08 mentioned, \textit{``When something feels abnormal, or when it stops for too long.''} P18 highlighted emergency braking, \textit{``If it suddenly brakes hard, I really hope it gives an explanation. Right now, there’s no voice or text. I’d like it to explain the situation, maybe why it adopted a certain avoidance strategy and why it had to brake frequently. I think it needs a more standardized explanation process.''} In contrast, others expressed a desire for explanations even in more routine scenarios, such as lane changes or acceleration (P17), suggesting that the vehicle could broadcast these actions similarly to navigation prompts.


Beyond in-vehicle driving behaviors, users also wanted explanations for unexpected events while waiting for the robotaxi. For example, P04 described, \textit{``An order was automatically canceled after waiting for half an hour without any explanation.''} 

Finally, users emphasized that explanations should not be purely informational but also provide emotional support. As P03 suggested, \textit{``Explanations could include some reassurance, not be so cold.''} P13 similarly noted that \textit{``Offering comfort can demonstrate a more human-centered and empathetic system design.''}





\textbf{Emergency control.} Users emphasized the need for appropriate user education regarding emergency controls, suggesting that the waiting period before pickup could be effectively used for this purpose. As P09 explained, \textit{``Before the person gets in the car, robotaxis should tell them what measures are in place to ensure their safety. That way, they’ll feel more at ease. For example, when you open the app, it could include an introduction that highlights a few key ways you can protect yourself.''} In addition, users expressed a strong preference for a physical emergency button that did not rely on network connectivity. As P08 noted, \textit{``I think there should be an emergency button. If something urgent happens, like an accident or extreme weather, there should be a button you can press to directly call for rescue. I want a real and physical button, because I worry that if there’s a network problem, I might not be able to send a request properly.''}

\textbf{System capability.} Users expected robotaxis to demonstrate stronger technological performance, particularly in high-precision localization. As discussed in Sec. \ref{PC-Flexibility}, the limited flexibility in parking behavior, such as fixed pickup and drop-off points, and inaccurate stopping locations, often led to inconvenience. In response, P14 suggested leveraging Bluetooth or GPS signals to precisely determine the user’s real-time location and dynamically select a suitable nearby stopping point, \textit{``In theory, it can detect my location through the Bluetooth signal from my phone, so when I’m nearby, it should be able to find a suitable place to stop. There are many devices emitting Bluetooth signals now, such as phones and earphones. But Bluetooth can be very unstable in places like airports or high-speed rail stations. So if it relies on Bluetooth sensing alone, the positioning may not be accurate enough at the moment. This probably needs further improvement. I think GPS-based positioning might be more precise.''}







\subsubsection{Experiential and affective expectations.}\label{ED-Experiential}

\textbf{Entertainment.} Users expected richer entertainment features and tended to regard the vehicle as an experiential space. They hoped that more experiential and interactive functions could be integrated into the ride. For example, P13 noted that \textit{``Because autonomous driving often involves traveling alone or only with friends, the vehicle becomes a relatively private environment. In this context, travel time could be better utilized by offering entertainment features, such as interactive games or playful activities during the ride.''} P1 hoped that \textit{``Robotaxi have the option to cast and watch television.''} 

\textbf{End-to-end automation.} Users expected their trips to be more effortless, with smoother automation that minimized both mental and operational load. Many robotaxi platforms relied on fixed pickup and drop-off points (as noted in Sec. \ref{PC-Flexibility}), which required users to manually adjust their locations to the nearby predefined stops, instead of their real locations. This design placed a usage burden on users. For instance, P09 stated that \textit{``I would prefer to simply enter my origin and destination and have the system automatically determine the most convenient pickup and drop-off points, rather than forcing users to make these judgments themselves. At present, I must manually locate available stations and decide which one is closest, which I find cumbersome.''} P09 further emphasized the desire for seamless integration between robotaxi services and public transit systems, \textit{``I compared my expectation to existing navigation apps that can plan multimodal journeys, combining walking, cycling, and public transportation into a single and coherent route. I hoped robotaxi services could be similarly integrated into the whole transportation system, such that the system would automatically plan where to transfer from the subway, where to walk, and where to board the robotaxi. Without such integration, using a robotaxi currently requires significant effort, including traveling into the vehicle’s operational zone, locating the correct pickup point, and figuring out the final destination stop. It's an extremely inconvenient and frustrating experience.''}

\subsubsection{Personalized and adaptable expectations.}\label{ED-personalization}
\textbf{Driving customization.} Driving customization was reflected in two main aspects. First, users wanted support for situational, ad hoc needs during a trip. Participants expressed a desire to make temporary routing or stopping requests, such as changing the destination or adding an intermediate stop, and expected the robotaxi to evaluate whether such requests could be accommodated and then confirm the outcome. For example, P10 described that \textit{``I want an AI-driven interaction in which the user could request a destination change or an additional waypoint, and the system would assess feasibility and respond accordingly.''}

Second, users emphasized the need for customizable driving modes. Some participants felt that current driving modes were overly simplistic and hoped for multiple predefined options, such as ``fast'' or ``casual'' modes. For instance, P08 suggested, \textit{``Robotaxis can have a special mode for urgent situations and another for more relaxed travel.''} Beyond speed alone, several users expected multidimensional customization, including factors such as ride smoothness, scenic quality of the route, travel time, and cost. P11 noted that \textit{``Some people might actually want to see more beautiful views and would be willing to take a longer and more indirect route for that.''} P11 highlighted the importance of time and money, \textit{``Some routes save time but cost more, for example, because the distance is longer. I hope there could be an option to choose. What I really care about is how long it takes and how much it costs.''}

Some users articulated more advanced expectations. Rather than selecting from predefined modes, they wanted to express situational intentions in natural language, which robotaxis would then translate into appropriate driving behavior. P10 suggested, \textit{``I think I could talk to it like this. ‘You can drive faster and don’t worry about my riding comfort. I’m in a hurry,' or `Right now I have an elderly passenger or a pregnant passenger, so please drive more smoothly. I’m not in a rush.' If I can make these kinds of requests and they can meet them, I think that would be really good.''} P08 directly envisioned an interactive planning process, \textit{``If I say I want to take a route with better scenery, it could generate Route 1, Route 2, and Route 3, and then I choose one. And not every passenger needs to state their preferences. It shouldn’t be like filling out a questionnaire before every ride. The system could simply show a pop-up asking whether there are any special needs.''}

Finally, one user imagined a more speculative future in which driving modes could be adjusted automatically without explicit user input, based on passive signals such as health monitoring data. P08 said that \textit{``If there were some kind of passenger health monitoring, I think that could be good. For example, using heart rate or blood pressure to adjust the driving mode. Of course, that would be with consent enabled.''}





\textbf{Personalized companion.}
One participant envisioned a robotaxi equipped with a caring in-vehicle virtual companion that provided a sense of companionship, with the communication style adapted to the user’s profile and context. P09 mentioned, \textit{``For example, for tourists, the companion could casually introduce scenery and local attractions in a natural and conversational tone. For children, it could engage in supportive and reflective dialogue to encourage expression and thinking, while for adolescents or adults, it could offer emotional support or simple comfort.''} 

When describing the companion’s personality, P09 emphasized adaptability as different users had different needs, envisioning that \textit{``Users who prefer minimal effort could select from preset options for the companion's personality, such as `knowledgeable' type or an `emotionally supportive' type, while others could set the companion’s style themselves.''}

Moreover, companionship extended beyond verbal interaction to coordinated in-cabin adjustments, as P09 noted,\textit{``During long or tiring journeys, the companion could help users relax by playing meditation music, guiding simple mindfulness exercises, adjusting lighting, music, or seat positions. These subtle and embodied interactions are accompanying the user in an intangible yet meaningful way.''}

Importantly, P09 expected this virtual companion to be tightly linked to the user’s personal account rather than the vehicle. When the user entered the car, their personalized companion would automatically activate. When they exited, the connection would be seamlessly released. She said, \textit{``It’s kind of like a DeepSeek account that’s trained only by you and follows your account. The companion set in the robotaxi account would be entirely your style. Once you get into the car, it connects to the system and talks to you based on the state of your phone.''}

\textbf{Generalized usage.} Several users envisioned a future in which robotaxis support a broader range of use cases. First, rather than functioning solely as point-to-point taxis, robotaxis were imagined as flexible and on-demand mobility resources. One anticipated scenario was a chartered robotaxi service, where users could book a vehicle for an extended period, such as an entire day. P10 said, \textit{``I could charter the car for the whole day. Wherever I want to go, it takes me there, and then it can leave on its own. When I need it again, it comes back to pick me up. That way, when I go out, I don’t have to drive myself or worry about finding parking, and I don’t need to think about things like a traditional rental driver, such as whether they’ve eaten or where they are. It feels completely hassle-free for me.''} In this model, the robotaxi eliminated the need to drive, find parking, manage a private vehicle, or coordinate with a human driver. 

Another mentioned use case was corporate or organizational deployment, including business reception and shuttle services. P10 mentioned that \textit{``Companies could rent multiple robotaxis to operate on fixed routes, such as between offices and airports or high-speed rail stations. This would replace traditional arrangements involving dedicated drivers and reception staff, enabling employees to accompany guests directly while the vehicle handles transportation autonomously. Besides, robotaxis could also be used as employee commuter shuttles operating on fixed daily routes.''}

Beyond transportation, some users also reconceptualized the robotaxi as a paid and mobile personal space rather than a pure commuting tool. For example, P05 expressed that \textit{``I hope it could let me rest in the car to stay there for a while without going anywhere. The car could just remain parked, and I could lie down or take a short break. I really like that feature. Maybe I’m tired, or it’s not that urgent, or I’m waiting for someone. If it lets me rest for half an hour or so, even with an extra charge, that would be fine.''}

\section{A USER-DRIVEN DESIGN FRAMEWORK FOR ROBOTAXIS}

\begin{figure}[htbp]
	\centering
		\includegraphics[scale=0.55]{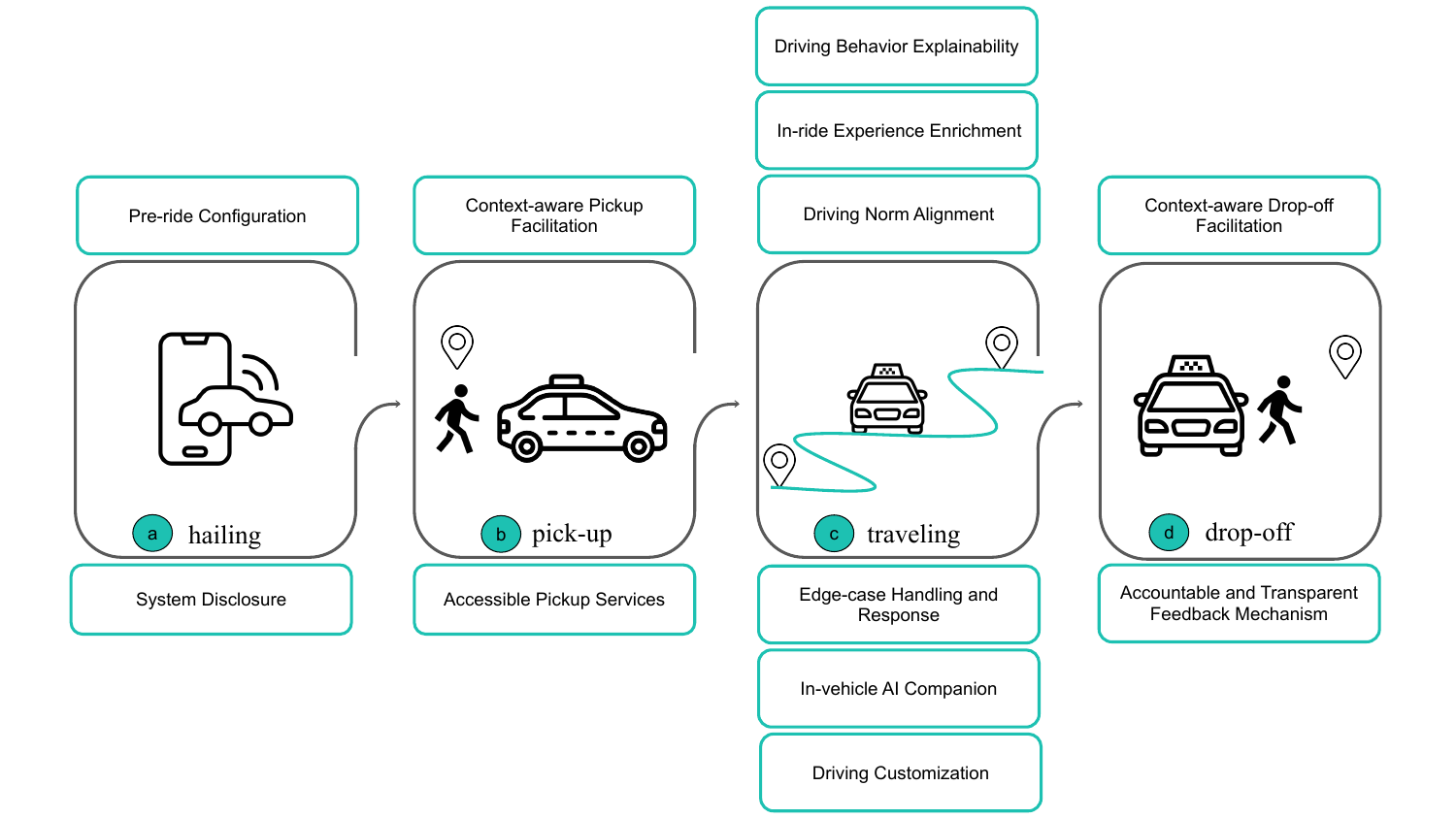}
	\caption{A User-driven Design Framework for Robotaxis.}
	\label{framework}
\end{figure}

Based on the findings, we develop a user-driven design framework for robotaxis spanning hailing, pick-up, traveling, and drop-off
phases, as shown in Figure \ref{framework}.

\subsection{Hailing and Before-Ride}

\subsubsection{Pre-ride Configuration}

As discussed in Sec. \ref{PA-agency}, a key advantage of robotaxis was enhanced user agency: users felt greater control when they could customize not only in-car settings such as temperature and music, but also external features like rooftop lighting. Furthermore, Sec. \ref{ED-personalization} shows that users expected personalization across a wide range of elements, from driving modes to companion personality. In this context, leveraging waiting time for \textit{pre-ride configuration} shows strong potential to reduce in-transit cognitive load, create anticipated gains, and provide a buffer for adjustments (e.g., temperature). Such pre-ride configurations support the autonomy and competence needs highlighted by Self-Determination Theory~\cite{deci2012self}.

\subsubsection{System Disclosure}

Sec. \ref{PC-Transparency} demonstrates that uncertainty in robotaxi services, spanning vehicle availability, arrival time, vehicle condition, and system status, could provoke user anxiety and erode trust. To mitigate these effects, we propose meaningful system disclosure at the hailing stage as a mechanism to enhance transparency. Such disclosure could be implemented across multiple levels: (1) Service disclosure, including supply availability and service coverage; (2) Trip predictive disclosure, such as estimated system response time, pickup wait time and distance, and travel price; (3) Vehicle and system disclosure, including real-time vehicle and system status; and (4) Safety and emergency disclosure, detailing emergency handling features and procedures. These forms of system disclosure align with prior research emphasizing high system transparency as a means of establishing user trust in robotaxi~\cite{liu2025understanding}.

\subsection{Pickup}

\subsubsection{Context-aware Pickup Facilitation}

As shown in Sec. \ref{PC-Flexibility}, limited pickup flexibility remained a key challenge. We propose \textit{context-aware pickup facilitation} to address this issue. We suggest that the system integrate multiple sensing signals to ensure that pickup points are both safe (e.g., avoiding puddles or heavy traffic) and convenient (e.g., reliably identifying hailing passengers through a combination of Bluetooth localization and visual perception~\cite{wu2025enhancing}). Beyond this, we call for intelligent communication channels that incorporate a virtual driver into pickup coordination, enabling fine-grained adjustments to location and waiting time.

\subsubsection{Accessible Pickup Services}

Sec. \ref{PC-Flexibility} further notes that robotaxis' lower flexibility in providing accessible pickup services, compared with human drivers, constituted a key limitation. To address this challenge, we propose \textit{accessible pickup services} as a key design direction, tailored to passengers with diverse needs. For instance, when the vehicle approaches, provide non-visual alerts (e.g., smartphone vibration or vehicle audible signals) to assist blind and low-vision passengers; when large luggage is detected, automatically open the trunk and provide luggage-loading assistance; when a passenger carrying an infant is identified, automatically open the door. These accessible pickup services are consistent with recent research advocating for more inclusive robotaxi design~\cite{goralzik2022shared,ma2025designing}.

\subsection{Traveling}

\subsubsection{Driving Behavior Explainability}

As discussed in Sec. \ref{ED-Control and trust expectations}, users regarded explainability as a core requirement of robotaxi systems, consistent with prior research~\cite{kuznietsov2024explainable}. Effective \textit{driving behavior explainability} should consider situational explanations, such as action–rationale explanations for decision transparency, anticipatory timing feedback for temporal predictability, and affective reassurance for emotional support. Importantly, explainability is not a case of ``more is better'', and users' needs and preferred modes of communication (e.g., when to provide explanations and whether users need emotional reassurance) vary considerably. Therefore, beyond conveying essential information (such as situational explanations during emergencies), explainability modules should support customization across multiple dimensions, such as communication triggers (e.g., event-driven vs. user-initiated), communication strategies (e.g., affective support vs. informational support), and communication modalities (e.g., auditory vs. visual).

\subsubsection{In-ride Experience Enrichment}

Our findings suggest that users anticipated expanded possibilities for entertainment and interaction (e.g., karaoke, streaming video, and gaming) during the ride in Sec. \ref{ED-Experiential}. 
In practice, robotaxis represent more than transportation, which constitute transitional spaces that can imbue fragmented travel time with greater experiential value~\cite{auhspace}. We therefore encourage designers and practitioners to broadly explore the potential of this transitional space, integrating activities such as relaxation, entertainment, and meditation into the robotaxi experience~\cite{auhspace,meurer2020wizard}.

\subsubsection{Driving Norm Alignment}

Most existing approaches to autonomous driving prioritize driving performance such as safety, efficiency, and rule compliance, with limited attention to human-aligned driving behavior~\cite{ma2025aln,fraade2024being,song2025drivecritic}. Sec. \ref{F-Ethics} highlights a critical ethical concern of users in robotaxi systems: the absence of human-like driving norms, which is often overlooked in Wizard-of-Oz studies. For example, robotaxis cannot send or interpret common social driving signals (e.g., flashing headlights to express thanks or using short horn signals on blind or S-shaped curves to warn oncoming traffic). As the integration of vision-language models into end-to-end autonomous driving systems becomes increasingly prevalent~\cite{song2025drivecritic,ma2025aln,tian2024drivevlm}, we call for future research to incorporate human driving norms into robotaxis' driving and communication, enabling robotaxis to better integrate into the broader transportation ecosystem.

\subsubsection{Edge-case Handling and Response}
Sec. \ref{PC-Robustness} identifies that users' concerns about robotaxi robustness, such as recognizing atypical road obstacles, adapting to diverse weather environments and solving network breakdown, might undermine user trust and willingness to use the system. Therefore, we recommend prioritizing edge-case training and real-world testing as key directions for performance improvement~\cite{rahmani2024systematic}. At the same time, effective in-vehicle and mobile-based information provision about edge-case handling could be used to support trust calibration~\cite{kraus2020more}. Moreover, the system could integrate AI and human oversight to enable efficient and accurate real-time monitoring of vehicle operation: when exceptional situations arise, the system must provide timely explanations and solutions, as well as support convenient and rapid vehicle disengagement options, such as clearly accessible physical control buttons~\cite{lim2021ux}.

\subsubsection{In-vehicle AI companion}

Although robotaxi companies such as Pony.ai have implemented a ``virtual driver'' with companion-like functionality~\cite{ponyai}, this study further reveals a broader and more nuanced set of expectations for \textit{in-vehicle AI companions}, as shown in Sec. \ref{ED-personalization}. In particular, users expressed a desire for adaptive conversational roles that responded to situational context, configurable companion styles that reflected individual preferences, embodied and multimodal forms of interaction, and account-bound companions that enabled continuity and long-term personalization across sessions. Generally, the ability to dynamically support users across informational, emotional, and decision-making dimensions is a critical direction for the design of robotaxi AI companions.

\subsubsection{Driving Customization}

This study captures users' expectations for \textit{driving customization} in robotaxis. Beyond basic needs such as ad hoc route changes or parking requests, users expressed a desire for customizable driving modes with natural language understanding, enabling preferences such as scenic- versus distance-optimized routing and efficiency- versus comfort-oriented driving styles, as discussed in Sec. \ref{ED-personalization}. Notably, driving customization blurs the boundary between machine and human decision-making and raises ethical concerns about responsibility attribution, placing greater demands on the ethics development of legal frameworks for autonomous driving~\cite{maurer2016autonomous,padinhakara2025navigating}.

\subsection{Drop-off and Post-ride}

\subsubsection{Context-aware Drop-off Facilitation}

Similar to pickup, limited flexibility in drop-off locations emerges as the primary inconvenience during the drop-off stage, as shown in Sec. \ref{PC-Flexibility}. To address this issue, we suggest context-aware drop-off facilitation as a solution. Inside the vehicle, this feature could promptly understand, confirm, and respond to users' drop-off intentions, such as voice commands like ``stop nearby'' or changes to the drop-off point made via the in-vehicle display or mobile interface. Outside the vehicle, it requires accurate recognition of surrounding drop-off conditions, including safety, accessibility, and overall convenience.

\subsubsection{Accountable and Transparent Feedback Mechanism}

As shown in Sec. \ref{F-Ethics}, users worried that when they were dissatisfied with a driving experience and provided feedback, it was unclear who was responsible and how that feedback would lead to improvement. In fact, how to meaningfully involve and incorporate lay users' feedback into algorithmic improvement has become a topic of concern in responsible AI~\cite{laine2024ethics,vincenzi2024lay,shen2021everyday}. This is especially critical for robotaxi systems, where AI decisions are directly tied to user safety~\cite{hong2020artificial}. Therefore, we call for robotaxi feedback systems to critically examine accountability mechanisms and transparency in feedback processing~\cite{lima2022conflict}. In particular, user feedback should not serve merely as a venting mechanism, but should instead be genuinely integrated into backend processes and accountability chains through structured feedback triage and remediation, with transparent visibility into how it is handled and acted upon~\cite{kinchin2024voiceless,sekwenz2025unfair}.


\section{DISCUSSION}

\subsection{Beyond Transportation: Robotaxis as an Autonomous, Semi-private Transition Space}

This paper examines how users construct meanings around the in-vehicle space of robotaxis, revealing it to be more than a site of transportation, but an \textit{autonomous, semi-private transition space} (Sec.~\ref{F-privacy}). First, the absence of a human driver reconfigures the experience of \textit{autonomy}: compared to driver-present vehicles, users perceived robotaxi monitoring systems as less socially intrusive, which afforded greater freedom to engage in personal activities. This perceived autonomy, however, does not imply that users have full control. Users remained aware of ongoing in-vehicle monitoring and modulated their behavior in response, with some explicitly acknowledging the benefits of such monitoring systems in maintaining safety and supporting performance improvement.

This dual awareness positions the robotaxi interior as neither fully private nor overtly surveilled. It cannot be equated with the intimate yet contested privacy intrusions associated with smart-home cameras~\cite{pierce2019smart,saqib2025bystander}, nor with the explicitly power-asymmetric and productivity-oriented surveillance characteristic of workplace settings~\cite{sum2025s}. Instead, robotaxis emerges as a \textit{semi-private space}, where autonomy and monitoring coexist through tacit negotiation and normalization.

In practice, robotaxis represent more than transportation. As users reconceptualize them as paid, mobile personal spaces, the in-vehicle environment functions as a \textit{transition space} that extends beyond movement between destinations (Sec. \ref{ED-personalization}). Rather than serving solely as a temporal gap in commuting, this space allows users to pause, wait, rest, or engage in self-directed activities, transforming fragmented travel time into a meaningful interval of personal use~\cite{auhspace}. Situated between mobility and stillness, and between public circulation and private retreat, the robotaxi interior supports experiential recalibration toward relaxation, entertainment, and reflection, pointing to broader opportunities for in-car experience design~\cite{auhspace,meurer2020wizard}.

However, users' trust in this semi-private space is contingent and carefully negotiated. They were sensitive to the power asymmetries embedded in in-vehicle data collection, and tended to focus their concerns on data use rather than data collection itself (Sec. \ref{F-privacy}). Questions about how privacy is protected, and how data may be accessed remain insufficiently transparent yet central to users' concerns (Sec. \ref{F-privacy}). Given the heterogeneity of data within the robotaxi ecosystem~\cite{al2024navigating,allessons}, which ranges from in-cabin audio and video recordings to vehicle location traces, and even personal information from mobile devices (e.g., Bluetooth identifiers or phone numbers), we call for future research to place greater emphasis on meaningful informed consent across different forms of data collection and use in robotaxi systems.

Notably, privacy perceptions in robotaxi systems are deeply shaped by social and cultural contexts. In China, public trust and support for in-vehicle data collection are often embedded within collectivist cultural values and governance narratives that prioritize order and safety~\cite{su2022explains}. Future work should therefore examine privacy perceptions and concerns across diverse cultural settings to inform privacy design in robotaxi ecosystems.

\subsection{Entrusting Safety to Machines: Toward Safe, Responsible, and Trustworthy Robotaxi Systems}


Our findings reveal nuances in the relationship between safety and ethics in robotaxis. On the one hand, replacing human drivers with fully automated decision-making offered potential safety benefits, such as more conservative and consistent driving styles, the elimination of risks from driver fatigue and road rage, and a reduced potential for conflict between passengers and drivers (Sec. \ref{F-Safety}). On the other hand, the delegation of passengers’ safety entirely to machines raised fundamental ethical questions, such as accountability and the ethical acceptability of such delegation, especially in the event of accidents (Sec. \ref{F-Ethics}). Our findings further reveal a critical gap between \textit{algorithmic safety} and \textit{perceived safety}. Users’ perceived safety was dynamically shaped through continuous sensemaking of vehicles’ driving decisions. As such, any counterintuitive decisions, however reasonable from the
system’s perspective, could amplify users’ uncertainty, distrust, and anxiety. In this respect, explainability may serve as a crucial bridge between system-level safety and users’ perceived safety by helping users make sense of vehicle behavior and underlying decision rationales. We therefore suggest that future research systematically explore the diverse dimensions of explainability design, such as thresholds and triggers for explanations~\cite{omeiza2021explanations}, emotional communication in explanations~\cite{li2024review}, and explanations in external communication~\cite{colley2022investigating}.


We also find that users, even without technical backgrounds, tended to actively make sense of the model development of robotaxis rather than treating them as opaque black boxes. Notably, users perceived multiple factors of uncertainty during sensemaking, from how developers translated driving experience into system design (Sec. \ref{F-Safety}), to how training data handled edge cases (Sec. \ref{PC-Robustness}). These uncertainty factors ultimately amplified users' concerns about machine decision-making. At the same time, many users used external information channels to obtain information about robotaxis (Sec. \ref{motivation}). On this note, designers and practitioners could investigate external information channels, such as press releases and social media discourse, to support informed understanding and trust calibration.


Finally, we identify the tension between passenger autonomy and accountability. While users desired customization of robotaxi driving modes, they expressed concern about liability in case of accidents following such customization. This concern echoes prior literature warning of moral and ethical ambiguity in fully autonomous driving~\cite{wang2025decision,li2016trolley,jia2024will}. As automated driving evolves from SAE Level 2 to Level 4, the human-vehicle collaboration paradigm shifts from machine-assisted humans, to human-assisted machines, and ultimately to human-customized machines. We call on legal scholars and practitioners to proactively consider how these evolving forms of human–vehicle collaboration should inform the development of autonomous driving legal frameworks.

\section{LIMITATIONS}\label{limitations}

Our study focused exclusively on robotaxi use in China, which might limit the generalizability of our findings to other cultural contexts. Future research could examine and compare user perspectives across cultures to clarify how cultural factors shape users' perceptions and usage behaviors in robotaxi systems. In addition, our interviewees were all young adults (i.e., approximately 20–30 years old) with relatively high educational levels (i.e., mainly bachelor's degrees). This sampling profile may underrepresent the needs and concerns of other demographic groups. Finally, the proposed user-driven design framework was developed from 18 interviews together with our researchers’ autoethnographic insights. As such, it should be viewed as an initial, user-centered set of design priorities rather than a comprehensive, all-encompassing framework. Future work with larger and more diverse samples and iterative validation in real-world deployments could further refine, extend, and systematize the framework.

\section{CONCLUSION}
Our study suggests that robotaxis were not experienced simply as a new ride-hailing option, but as an autonomous, semi-private transitional space that reshapes the human-vehicle collaboration paradigm. Through interviews and autoethnographic rides, we found that users valued a range of benefits (e.g., the agency and consistency robotaxis can offer), yet also encountered persistent frictions (e.g., limited flexibility and insufficient transparency), alongside broader negotiations around privacy, safety, ethics, and trust. To support more socially acceptable deployment, we offer a user-driven, end-to-end design framework that translates these insights into actionable directions for trustworthy, transparent, and accountable robotaxi design.

\bibliographystyle{ACM-Reference-Format}
\bibliography{sample-base}

\appendix
\newpage
\section{Interview Protocol}\label{interview protocol}

\begin{enumerate}

\item \textbf{Participant Background}

\begin{enumerate}
    \item Could you share your age, gender (optional), education level, and the city you live in?
\end{enumerate}

\item \textbf{General Experience with Taxis and Ride-hailing}
\begin{enumerate}
    \item About how many times do you take a taxi or ride-hailing service in a typical month?
    \item How do you usually split your trips across different modes (street-hailed taxis, ride-hailing services with a driver, and robotaxis)?
    \begin{enumerate}
        \item Roughly how often do you use each?
        \item In what situations would you choose street-hailed taxis vs. ride-hailing vs. robotaxis, and why?
    \end{enumerate}
    \item Which apps do you use most frequently?
\end{enumerate}

\item \textbf{General Discussion About Robotaxis}

\begin{enumerate}
    \item How many times have you used a robotaxi?
    \item When was the most recent time you took a robotaxi?
    \item Which city have you used robotaxis in?
    \item Which app or platform did you use to request the robotaxi?
    \item When do you typically use robotaxis (time of day)?
    \item What were the conditions usually like when you rode robotaxis?
    \begin{enumerate}
        \item How was the weather?
        \item How was the traffic?
        \item How was the surrounding environment (e.g., downtown, residential, campus, business district)?
    \end{enumerate}
\end{enumerate}
\item \textbf{Motivations for Using Robotaxi}
\begin{enumerate}
    \item What motivated you to take your first robotaxi ride?
    \item When did you first use a robotaxi, and why did you decide to try it at that time?
    \item In what scenarios do you typically choose robotaxis now?
    \item Why do you choose a robotaxi instead of other transportation options in those situations?
\end{enumerate}

\item \textbf{Perceived Benefits and Challenges}
\begin{enumerate}
    \item Overall, how do you feel about robotaxi services?
    \begin{enumerate}
        \item What do you think are the good parts?
        \item What do you think are the not-so-good parts?
    \end{enumerate}
    \item Before using a robotaxi, did you have any expectations or concerns? What were they?
    \item After using robotaxis, did your perception change in any way? How?
    \item Compared with traditional ride-hailing or street taxis, what do you see as the main advantages and disadvantages of robotaxis?
    \item Have you had any moments that were especially memorable (either positive or negative)?
    \item Have you ever had a particularly bad robotaxi experience? What happened?
\end{enumerate}
\item \textbf{Safety Perceptions}
\begin{enumerate}
    \item Do you think robotaxis are safe? Why or why not?
    \item Which feels safer to you: a street-hailed taxi, a ride-hailing car with a human driver, or a robotaxi? Why?
    \item Have you ever experienced a ``close call'' or a moment when you felt nervous during a robotaxi ride?
    \begin{enumerate}
        \item What was the situation (location, approximate speed, weather, surrounding traffic)?
        \item What triggered the risk or discomfort (e.g., pedestrians, another car cutting in, construction)?
        \item Did the system provide any warning or prompt?
        \item How did you react in that moment?
        \item Was there any human intervention (e.g., remote support), or did you press any in-car/app button?
        \item After the incident, did your opinion of robotaxis change? If so, how?
    \end{enumerate}
\end{enumerate}
\item \textbf{Privacy Perceptions}
\begin{enumerate}
    \item Do you feel you have privacy in a robotaxi? Why or why not?
    \item Compared to rides with a human driver, do you feel more worried or less worried about privacy in a robotaxi? Why?
    \item What privacy concerns do you have when using robotaxis?
    \item What types of data do you think a robotaxi service might collect about you?
\end{enumerate}
\item \textbf{Trust Perceptions}
\begin{enumerate}
    \item Overall, do you trust robotaxis? Why or why not?
    \item How much do you trust robotaxis at different stages of the journey? (hailing, pick-up, traveling, drop-off)?
    \item Do you trust robotaxis to drive safely? Why or why not?
    \item Do you trust robotaxis to drive accurately in terms of routing and navigation (e.g., taking detours, finding you, choosing pickup/drop-off points)? Why or why not?
    \item In what situations do you trust robotaxis more, and in what situations do you trust them less? (e.g., highways, main roads, small streets, nighttime, rain/snow, complex intersections, campus) Why?
\end{enumerate}
\item \textbf{Design Expectations (Following the Journey Walkthrough)}

\begin{enumerate}
    \item \textbf{Hailing}
    \begin{enumerate}
        \item How did you place the order?
        \item When requesting a robotaxi, what are you most worried about?
        \item While waiting, what app/system messages do you receive or pay attention to?
        \item In a ride with a human driver, what would you typically contact the driver about before pickup?
        \item Would you want robotaxis to provide an equivalent way to handle those needs? If yes, what should it look like?
         \item What support would you expect from the robotaxi during the ordering and waiting stage?
    \end{enumerate}
    \item \textbf{Pick-up}
    \begin{enumerate}
        \item How does the system notify you that the vehicle has arrived?
        \item How do you confirm that ``this is actually my vehicle''?
        \item Do you need to do anything to start the ride once you get in?
        \item Was the boarding process smooth? Did anything make you nervous or confused?
        \item Have you ever failed to find the vehicle or experienced a timeout? What reminders or alerts did you receive in that situation?
        \item Are there any special in-vehicle devices or interfaces (e.g., screens, voice system)? How do you use them?

        \item With a human driver, what would you normally communicate during pickup?
        \item Would you want robotaxis to provide an equivalent interaction or support? If yes, how?
        \item What support would you expect from the robotaxi for confirmation and pick-up?
    \end{enumerate}
    \item \textbf{Traveling}
    \begin{enumerate}
        \item What do you usually do during the ride (e.g., listen to music, observe the road)?
        \item What signals do you pay attention to most (in-car screen, voice prompts, vehicle movements)?
        \item Has the robotaxi ever done something you did not expect? What happened?
        \item Does the robotaxi behave the way you would expect in situations like lane changes, overtaking, or construction zones? Why or why not?
        \item Have there been any unexpected events (e.g., sudden braking, detours)? Did the robotaxi explain the behavior? 
        \item Have you ever wanted to change the route or drop-off point during the ride? How did you communicate or handle that?

        \item With a human driver, what would you typically communicate while en route?

        \item Would you want robotaxis to provide an equivalent way to communicate or adjust the ride? What would that be?

        \item What support would you expect from the robotaxi during the ride?

    \end{enumerate}
    \item \textbf{Drop-off}
    \begin{enumerate}
        \item After arriving, do you need to do anything to end the trip?
        \item Did the drop-off location meet your expectations? If not, how would you provide feedback or resolve it?
        \item With a human driver, what do you usually do after arrival (e.g., rating, feedback, complaints)?

        \item Would you want robotaxis to provide similar post-trip options? If yes, what specifically?
        
        \item What support would you expect from the robotaxi during the drop-off stage?
        
    \end{enumerate}
    \item Is there anything important I didn’t ask that you think we should include?
\end{enumerate}

\end{enumerate}

\end{document}